\documentclass[journal]{IEEEtran}

\usepackage{float} \restylefloat{figure} 
\usepackage{color} 

\usepackage{graphicx}
\graphicspath{{eps/}}

\usepackage{rotating}
\usepackage{lscape}
\usepackage{bm} 
\usepackage{cite}
\usepackage{url}
\usepackage[small]{caption}
\usepackage{subfig} 
\usepackage{array} 

\usepackage{amsmath}   
\usepackage{afterpage}
\long\def\symbolfootnote[#1]#2{\begingroup\def\thefootnote{\fnsymbol{footnote}}
\footnote[#1]{#2}\endgroup}
\DeclareCaptionType{copyrightbox}
\usepackage{epstopdf}
\begin{document}

\title{{\Large \textbf{Triplet Spike Time Dependent Plasticity: A floating-gate Implementation}}}

\author{Roshan Gopalakrishnan,~\IEEEmembership{Student Member,~IEEE} and Arindam Basu,~\IEEEmembership{Member,~IEEE}
\thanks{Roshan Gopalakrishnan and Arindam Basu are with VIRTUS, IC design centre of excellence, School of Electrical and Electronic Engineering, Nanyang Technological University, Singapore 639798 (e-mail:arindam.basu@ntu.edu.sg).}
\thanks{Copyright (c) 2010 IEEE. Personal use of this material is permitted. However, permission to use this material for any other purposes must be obtained from the IEEE by sending an email to pubs-permissions@ieee.org.} }

\maketitle
\begin{abstract}
Synapse plays an important role of learning in a neural network; the learning rules which modify the synaptic strength based on the timing difference between the pre- and postsynaptic spike occurrence is termed as Spike Time Dependent Plasticity (STDP). The most commonly used rule posits weight change based on time difference between one presynaptic spike and one postsynaptic spike and is hence termed doublet STDP (D-STDP). However, D-STDP could not reproduce results of many biological experiments; a triplet STDP (T-STDP) that considers triplets of spikes as the fundamental unit has been proposed recently to explain these observations. This paper describes the compact implementation of a synapse using single floating-gate (FG) transistor that can store a weight in a nonvolatile manner and demonstrate the triplet STDP (T-STDP) learning rule by modifying drain voltages according to triplets of spikes. We describe a mathematical procedure to obtain control voltages for the FG device for T-STDP and also show measurement results from a FG synapse fabricated in TSMC 0.35$\mu$m CMOS process to support the theory. Possible VLSI implementation of drain voltage waveform generator circuits are also presented with simulation results.
\end{abstract}


\IEEEpeerreviewmaketitle

\section{Introduction}

Over the past ten years, numerous experimental studies \cite{Markram1997, bi_poo, zhang, stdp_abbott} have shown that the synaptic strength varies as a function of the precise spike timing difference $\Delta t = t_{post} - t_{pre}$ between the firing times $t_{pre}$ and $t_{post}$ of the presynaptic and postsynaptic neurons respectively. This synaptic plasticity rule, called Spike Time-Dependent Plasticity (STDP), has evolved as one of several unsupervised plasticity rules that play an important role in learning and memory in the brain. The mathematical model of STDP based on a pair of pre- and post-synaptic spike is referred as doublet STDP (D-STDP) while the one based on triplet of synaptic spikes
\cite{Bi_Wang_2002, Froemke_Dan_2002, wang, Froemke2006 } i.e either pre-post-pre synaptic spike or post-pre-post synaptic spike is referred to as triplet STDP (T-STDP). Several variants of these rules have also been proposed for pattern classification tasks \cite{SWAT, Mitra2009}.

 Experimental results \cite{pfister_triplet} indicate that D-STDP model based on pairs of spikes are not sufficient to explain synaptic changes due to triplets or quadruplets of spikes. D-STDP model also fails to reproduce frequency effects. However, T-STDP model can reproduce frequency effects along with the explanation of synaptic changes due to triplets and quadruplets of spikes. It is also important to be able to replicate rate based plasticity experiments. It has been demonstrated that the Bienenstock-Cooper-Munro (BCM) learning rule \cite{BCM} based on firing rates can be obtained from D-STDP when pre-synaptic and post-synaptic neurons fire uncorrelated or weakly correlated Poisson spike trains, and only nearest-neighbor spike interactions are taken into account \cite{Izhikevich_lettercommunicated}. However, it is not possible to make a strict theoretical mapping from the nearest-spike interactions D-STDP models to the BCM rule \cite{pfister_triplet} whereas T-STDP allows for such theoretical mapping. Apart from these experimentally observed benefits of T-STDP model, there are some computational advantages as well\cite{Gjorgjieva}. It has been shown that T-STDP can detect input correlations higher than the second order ones to which D-STDP is sensitive. Hence, it has been shown to drive direction and orientation selectivity\cite{Gjorgjieva}. Further, it can be shown to reproduce a more generalized version of BCM overcoming the limitations of the original one. Though there are several models of plasticity with varying degrees of bio-realism \cite{Mayr_algo}, T-STDP is a good compromise between simplicity and richness of function. Its simplicity also lends itself to easy analysis making it a good choice for a plasticity rule with functionality beyond D-STDP.

Recently STDP became so popular in computational neuroscience that neuromorphic engineers who try to emulate brain function using VLSI have also tried to emulate this behaviour in silicon. However, implementing a compact learning synapse continues to be one of the big challenges in the field \cite{Bo_Hasler}. Several recent papers have reported D-STDP implementations \cite{arthur_gamma, giacomo_spikestdp, dudek-2012, Koickal_Tara, STDP_Mems} and T-STDP implementation \cite{mostafa_triple_2013}; however, these synapses could either only store states in a transient fashion (using charge on a capacitor) or only hold two states in the long term. The size of these synapses are also large hindering scalability of these designs. A promising solution for non-volatile analog weight storage in very small area is provided by a floating-gate (FG) device \cite{Bo_Hasler, brink_learningfg, FG_STDP_Pankaala, Roshan_IJCNN2014, Roshan_TNNLS2015, haas, liu_iscas_syn, Smith2014, fg_stdp}. This concept was utilized recently to show weight storage and adaptation due to quantum phenomena based on input signal timing \cite{fg_stdp}.

Compared to other work, here we demonstrate for the first time the implementation of the T-STDP rule in a FG synapse by appropriately modifying the drain voltage pulse based on spike triplets. From chip measurement results, we show that FG synapse in Fig. \ref{fig:FG_mux_tun}(b) can reproduce (1) D-STDP learning window and (2) T-STDP results when appropriate control signals are applied on its terminals. Some initial results for triplet experiment based on this work were presented in \cite{Roshan_ISCAS2015}. In this paper, we present results for quadruplet experiments and frequency effects as well as a new drain voltage generation scheme. We also present circuits for VLSI implementation of the drain voltage waveforms.

The paper is organized as follows: Section \ref{sec:Stdp synaptic modification rule} provides a brief explanation of STDP models, D-STDP and T-STDP. Section \ref{sec:fg_basics} introduces the working of floating gate synapse, relation between FG synapse and D-STDP rule and then, finally arrive at the relation between FG synapse phenomenon and T-STDP rule. The experimental results are included in section \ref{sec:result}. Section \ref{sec:drain} presents a hardware implementation of the drain voltage waveform generator along with the simulation results. Finally the paper is concluded with discussions on consolidated work.

\section{STDP Synaptic Modification Rule}
\label{sec:Stdp synaptic modification rule}
In biology, synapses are specialized structures that permit the transfer of signals between two neurons with an associated synaptic strength or weight.
Learning typically implies the modification of synaptic weight due to the activities of the pre- and post-synaptic neurons. The STDP models are explained in detail next.

\subsection{Doublet STDP (D-STDP) Model}
\label{sec:D-STDP}

\begin{figure}	\centering
		\includegraphics[scale=0.8]{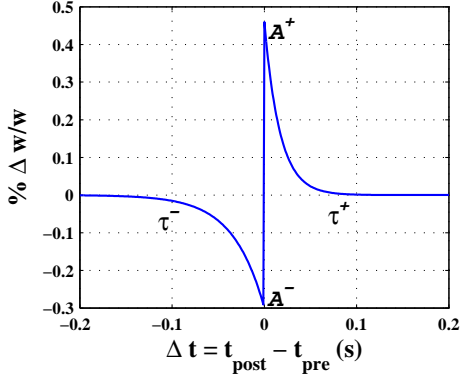}
\caption{STDP learning window; theoretical implementation of D-STDP: The plot is based on equation (\ref{eq:doublet_stdp}) with the parameters $A^+$ = $4.6\times 10^{-3}$ and $A^-$ = $3\times 10^{-3}$}.
	\label{fig:STDP_pfister}
\end{figure}

In D-STDP, potentiation occurs when a postsynaptic spike succeeds a presynaptic spike; otherwise depression happens. The weight changes can be governed by a temporal learning window. The temporal learning window for  STDP can be expressed as \cite{mostafa_triple_2013, pfister_triplet}
\begin{equation}
\begin{aligned}
	\label{eq:doublet_stdp}
	\Delta w=
	\begin{cases}
	\Delta w^{+}=A^+ e^{(\frac{-\Delta t}{\tau_+})} & if \quad \Delta t \geq 0 \\
	\Delta w^{-}=-A^- e^{(\frac{\Delta t}{\tau_-})} & if \quad \Delta t < 0 \\
	\end {cases}
\end{aligned}
	\end{equation}
where $\Delta t = t_{post} - t_{pre}$ is the time difference between a post-synaptic and pre-synaptic spike, $\tau_+$ and $\tau_-$ are the time constants of the learning window, and $A^+$ and $A^-$ are the maximal weight changes for potentiation and depression, respectively.
The theoretical graph for the above equation is simulated using MATLAB and is shown in Fig. \ref{fig:STDP_pfister} with the parameters being obtained by data fitting as explained in \cite{pfister_triplet}. As mentioned in \cite{bi_poo}, $\tau_+$ and $\tau_-$ are taken as $16.8$ms and $33.7$ms respectively for the simulation.

\subsection{Triplet STDP (T-STDP) Model}
\label{sec:T-STDP}

Previous studies \cite{sjostrom, wang}  show that the D-STDP model fails to reproduce the experimental outcomes involving higher order spike patterns such as triplet and quadruplets of spikes and furthermore, fails to account for the observed weight dependence on repetition frequency of pairs of spikes. To resolve the above mentioned issues, the D-STDP model was extended in \cite{pfister_triplet} to include spike triplets resulting in T-STDP model which could sufficiently reproduce physiological experiments.

The T-STDP rule is written as a function of difference in spike timings as, \cite{mostafa_triple_2013,pfister_triplet}
\begin{equation}
\begin{aligned}
	\label{eq:triplet_stdp}
	\Delta w^{+}= e^{(\frac{-\Delta t_1}{\tau_+})} (A_2^+ + A_3^+ e^{(\frac{-\Delta t_2}{\tau_y})}) \quad if \quad t = t_{post} \\
	\Delta w^{-}=- e^{(\frac{\Delta t_1}{\tau_-})} (A_2^- + A_3^- e^{(\frac{-\Delta t_3}{\tau_x})}) \quad if \quad t = t_{pre}
\end{aligned}
	\end{equation}
 where $A_2^+$  and $A_2^-$ denote the amplitude of the weight change whenever there is a pre-post pair or a post-pre pair respectively. Similarly, $A_3^+$  and $A_3^-$ denote the amplitude of the triplet term for potentiation and depression, respectively.

\begin{figure}	\centering
	\includegraphics[width=7cm, height=6cm]{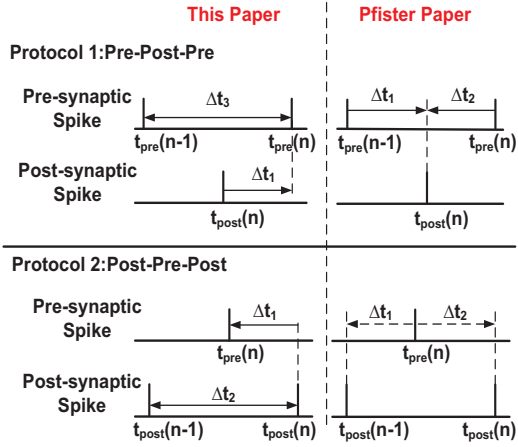}
    \caption{Timing diagram of T-STDP rule: The timing specifications of data points (x-axis) for triplet experiments: protocol $1$ and $2$. A comparison of the temporal notations in this paper and \cite{pfister_triplet} (mentioned as Pfister Paper in figure) is shown.}
    \label{fig:T-STDP_timing_dia}
\end{figure}

$\Delta t_1$ = $t_{post}$(n) - $t_{pre}$(n) , $\Delta t_2 $= $t_{post}$(n) - $t_{post}$(n-1) and $\Delta t_3 $= $t_{pre}$(n) - $t_{pre}$(n-1) are time difference between combinations of pre and post-synaptic spikes as shown in Fig. \ref{fig:T-STDP_timing_dia}. $\tau_- , \tau_+ ,\tau_x$ and $\tau_y$ are time constants for the above spike pairings.

In \cite{pfister_triplet}, though the T-STDP rule above is introduced first, it is shown later that not all terms are needed to explain biological data. Thus two different minimal models are defined later: (1) $A_2^+$ = $0$  and $A_3^-$ = $0$ for visual cortex data and (2) $A_3^-$ = $0$ for hippocampal culture data set. Hippocampal culture data set \cite{wang} is used for obtaining the results for triplets of spikes whereas visual cortex data \cite{sjostrom} is used for showing the frequency effects of T-STDP rule. For visual cortex data set, equation (\ref{eq:triplet_stdp}) simplifies to,
\begin{equation}
\label{eq:triplet_stdp_vc}
\Delta w(t)=
\begin{cases}
\Delta w^{+}= A_3^+ e^{(\frac{-\Delta t_2}{\tau_y})} e^{(\frac{-\Delta t_1}{\tau_+})}; \quad t = t_{post}    \\
\Delta w^{-}=- e^{(\frac{\Delta t_1}{\tau_-})} (A_2^-);  \quad t = t_{pre}
\end {cases}
\end{equation}
On the other hand, for hippocampal culture data set, equation (\ref{eq:triplet_stdp}) simplifies to,
\begin{equation}
	\label{eq:triplet_stdp_hippo}
	\Delta w(t)=
	\begin{cases}
	\Delta w^{+}= e^{(\frac{-\Delta t_1}{\tau_+})} (A_2^+ + A_3^+ e^{(\frac{-\Delta t_2}{\tau_y})}); \quad t = t_{post}    \\
	\Delta w^{-}=- e^{(\frac{\Delta t_1}{\tau_-})} (A_2^-);  \quad t = t_{pre}
	\end {cases}
	\end{equation}
For both cases, $\Delta w^{-}$ is exactly same as the case of long term depression (LTD) in D-STDP as shown in equation (\ref{eq:doublet_stdp}). Hence, to implement the triplet rule in circuits, we only need to modify pre-existing FG design to add the extra term in the potentiation case.

\section{Floating Gate Synapse}
\label{sec:fg_basics}

\begin{figure}	\centering
   	\begin{minipage}[b]{0.5\textwidth}
        \centering
		\includegraphics[width=7cm, height=6cm] {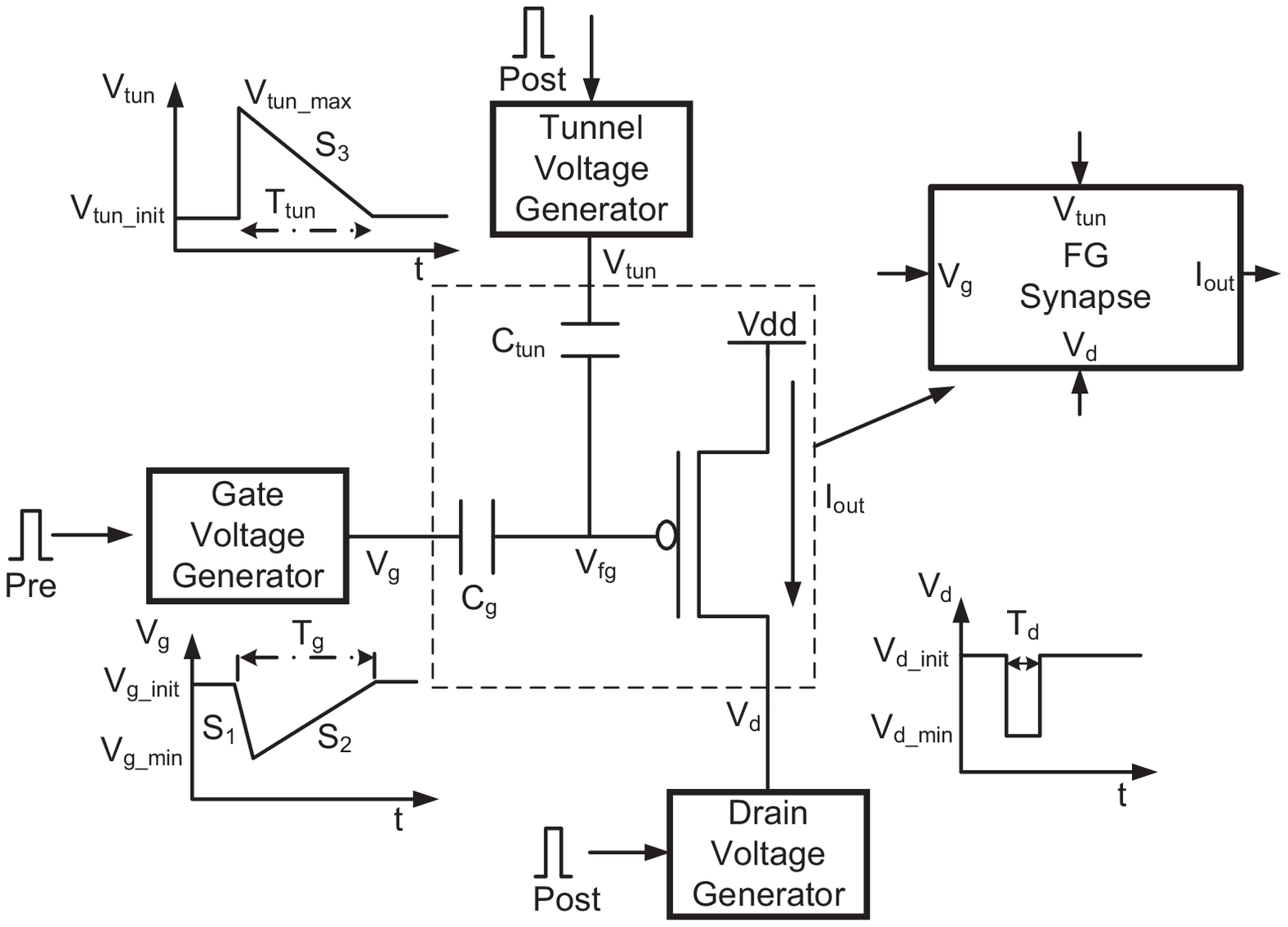} \\ (a) \\
 	\end{minipage}
   	 \begin{minipage}[b]{0.5\textwidth}
        \centering
		\includegraphics[width=7cm, height=6cm] {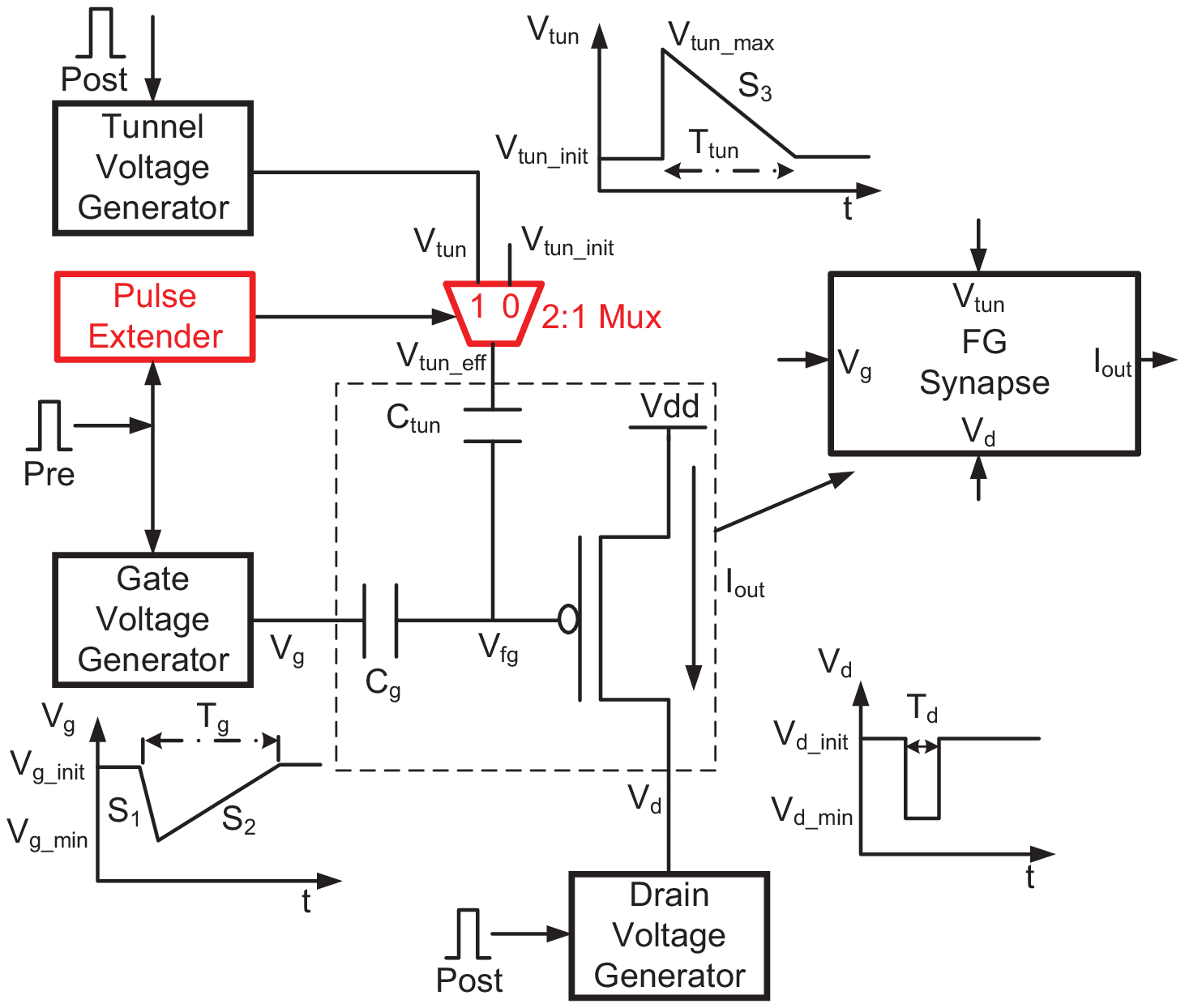} \\ (b) \\
    	\end{minipage}
    \caption{Floating Gate (FG) synapse: (a) The architecture used in previous work. (b) The architecture of a FG synapse used in this work with different input terminal voltage waveforms. The additional blocks used in the architecture is shown in red.}
    \label{fig:FG_mux_tun}
    \end{figure}

\begin{figure*}	\centering
    \begin{minipage}[b]{0.3\textwidth}
        \centering
		\includegraphics[width=6cm, height=5.5cm] {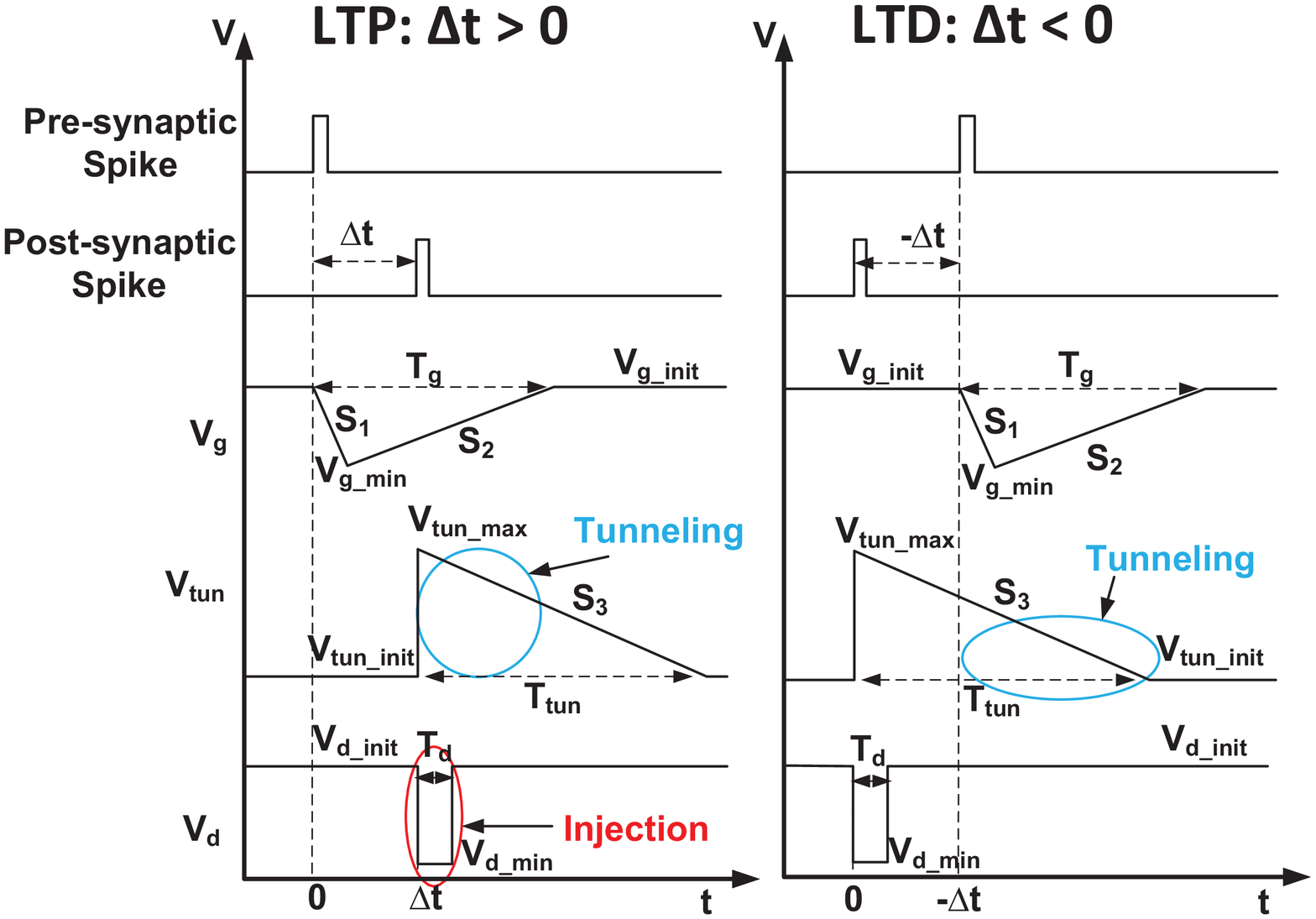} \\ (a) \\
	 \end{minipage}
    \begin{minipage}[b]{0.68\textwidth}
        \centering
		\includegraphics[width=11cm, height=5.5cm] {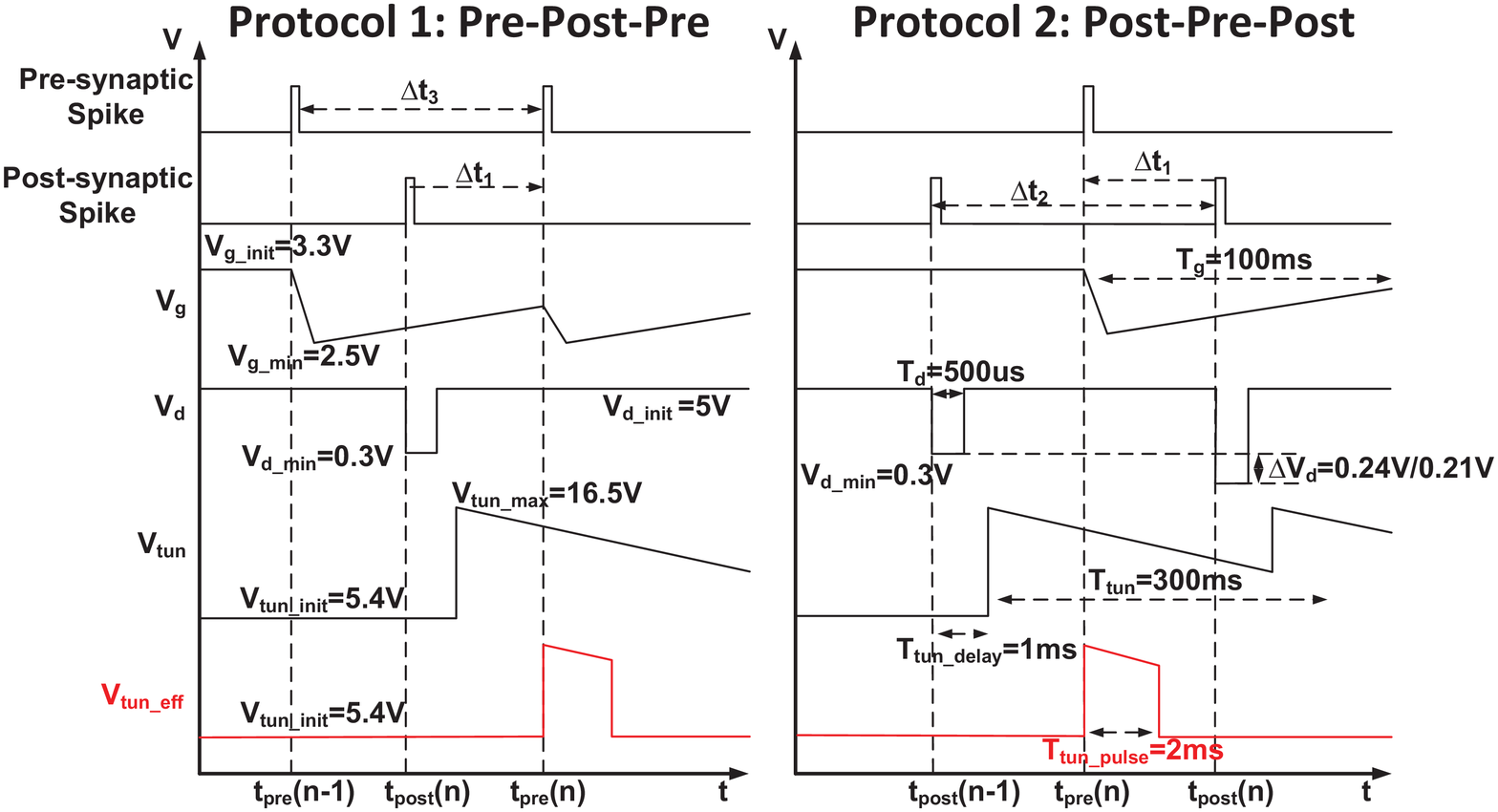} \\ (b) \\
    \end{minipage}
    \begin{minipage}[b]{1\textwidth}
        \centering
		\includegraphics[width=18cm, height=7cm]{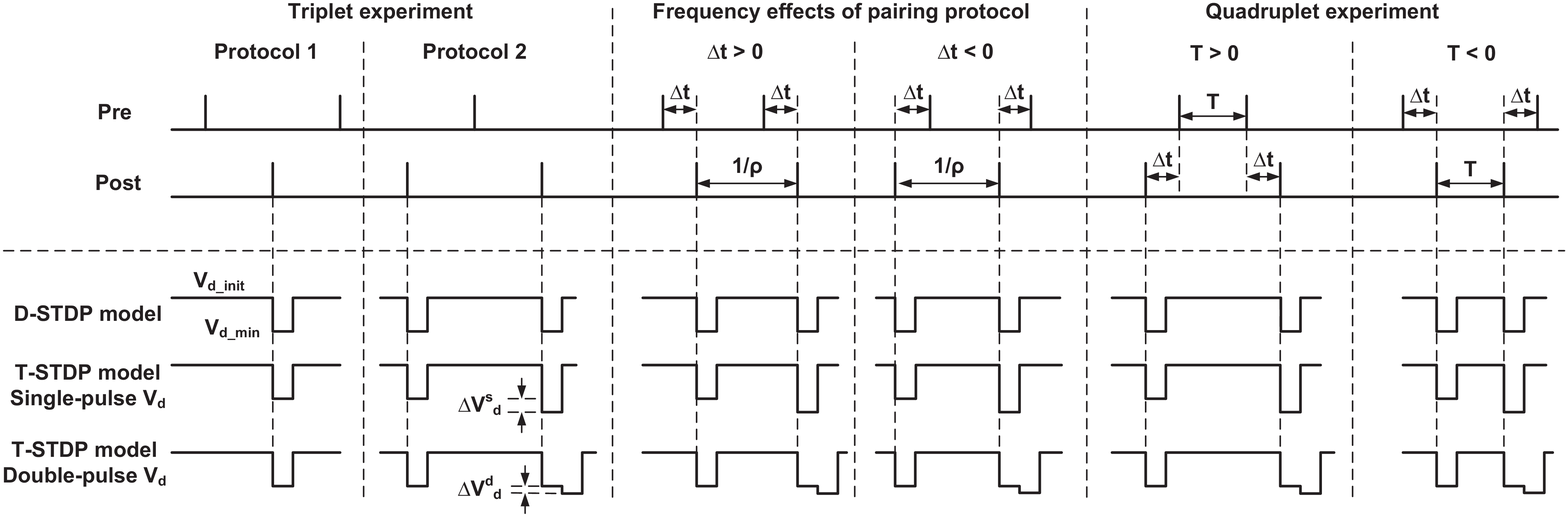} \\(c) \\
    \end{minipage}
        \caption{Timing diagram and voltage specifications of FG synapse: (a) D-STDP model, cases of long term potentiation (LTP) and long term depression (LTD), where effect of injection and tunneling is present in LTP where as only tunneling effect is present for LTD. These waveforms are used in our previous paper (b) The details of timing and voltage specifications used for testing the FG synapse for both triplet protocols is shown here. The modification of tunneling voltage, $V_{tun\_eff}$ (shown in red) sampled from $V_{tun}$ is shown. (c) Different cases of drain voltage waveform, $V_d$ for different experiments like pairing protocol, triplets and quadruplets.}
        \label{fig:STDP_timing_dia}
\end{figure*}

Fig. \ref{fig:FG_mux_tun}(a) shows the architecture of a single floating gate synapse in prior work \cite{fg_stdp}. It has three main terminals for programming as shown in the dashed box. The terminals are named as gate, drain and tunnel terminals with the respective voltages denoted as $V_g$, $V_d$ and $V_{tun}$. A ``non-STDP" behaviour seen in this work is ameliorated in our previous work \cite{Roshan_IJCNN2014, Roshan_TNNLS2015} along with detailed analysis of the operation of the D-STDP learning rule in a floating gate synapse. In previous works \cite{fg_stdp, Roshan_IJCNN2014, Roshan_TNNLS2015} , the quantum mechanism of tunneling is spread across a larger time scale (illustrated in Fig. \ref{fig:STDP_timing_dia}(a) and (b)) which makes it difficult to analyze the effect of triplets and quadruplets of spikes in a floating gate synapse. Similar to an approach in \cite{liu_iscas_syn, Smith2014}, the effect of tunneling can be localized at the occurrence of pre-synaptic spikes with the modification (red blocks) shown in the architecture of Fig. \ref{fig:FG_mux_tun}(b) making it easier to mathematically analyze the weight change for triplets. In the new architecture, whenever a pre-synaptic spike occurs, a triangular gate voltage waveform is generated which will create an exponential excitatory post-synaptic current (EPSC), similar to biology, because of the exponential relationship between the gate voltage and drain current of the MOS transistor in subthreshold region. The current at the maximum gate voltage is nearly zero. Similarly, whenever a post-synaptic spike arrives, a global triangular tunnel voltage waveform and an inverted pulse drain voltage waveform is generated. The global triangular tunnel voltage waveform is then sampled at the occurrence of pre-synaptic spike with the help of pulse extender block and the multiplexer. This creates a voltage waveform, $V_{tun\_eff}$ at the tunnel terminal. Thus, in the new architecture, during pre-synaptic spike, a gate voltage, $V_g$ and a tunnel voltage, $V_{tun\_eff}$ waveforms are generated and during post-synaptic spike, a drain voltage, $V_d$ waveform is generated. Whereas, in the previous architecture, a gate voltage waveform is generated during pre-synaptic spike and both drain voltage and tunnel voltage waveforms are generated during post-synaptic spike.

The equation for drain current of a subthreshold saturated pFET whose well is tied to $V_{dd}$ is given by \cite{fg_stdp}
\begin{equation}
\begin{aligned}
	\label{eq:FG_drain_current}
    I_{d}=I_{s0}e^{\kappa (V_{dd}-V_{fg})/U_{T}}
\end{aligned}
\end{equation}
where $U_{T}$ is the thermal voltage and $\kappa$ is the gate coupling coefficient.

Weight modification in a FG synapse uses a combination of hot-electron injection (HEI) and Fowler-Nordheim tunneling \cite{fg_stdp}. HEI adds electrons on to the floating gate node, which reduces the floating gate voltage resulting in more current through the transistor hence increasing the weight of the synapse. On the other hand tunneling removes electrons from the FG node to reduce the synaptic weight. In our previous paper \cite{Roshan_IJCNN2014}, for the case of D-STDP, a gate voltage waveform is generated at every pre-synaptic spike while at every post-synaptic spike, a tunneling voltage and a drain voltage is generated as illustrated in Fig. \ref{fig:STDP_timing_dia}(a). This shows that at every pre-synaptic spike only tunneling happens while at every post-synaptic spike there is both injection and tunneling. 
The occurrence of both tunneling and injection at the post-synaptic spike arrival necessitates the effect of injection to be greater than tunneling 
to obtain potentiation as in the traditional D-STDP learning window. Moreover since the effect of tunneling is spread over a long time, mathematical analysis or intuitive understanding of the effect due to multiple pre- and post-synaptic spikes becomes difficult. In contrast, the mathematical versions of both STDP models have potentiation and depression of weights localized at pre- and post-synaptic events. Hence, to relate the mathematical analysis of FG synapse with the T-STDP model easily, we also decided to localize the effect of tunneling and injection at pre- and post-synaptic pulses respectively. A similar technique has also been used in \cite{liu_iscas_syn, Smith2014}.

An illustration for this is shown in Fig. \ref{fig:STDP_timing_dia}(b) where a tunneling voltage waveform is still created at every post synaptic pulse as in earlier work \cite{Roshan_IJCNN2014, fg_stdp} but is not applied directly to the tunneling junction of the FG. Instead, it is sampled by the pre-synaptic pulse through a multiplexer to create a new waveform $V_{tun\_eff}$, which is applied to the FG (refer to Fig. \ref{fig:FG_mux_tun}). Figure \ref{fig:STDP_timing_dia}(b) also shows the specifications of terminal voltage waveforms for the case of triplets of spike. The well for the high voltage PMOS in the multiplexer can be shared with tunneling junction; nevertheless, this incurs an area penalty motivating us to look at possibilities of using non-localized tunneling for T-STDP in future.
The governing equations for injection and tunneling are given as \cite{fg_stdp},\cite{basu_iscas}
\begin{equation}
\begin{aligned}
	\label{eq:FG_phenomenon}
    & I_{inj}=I_{inj0} (I_{d}/I_{s0})^{\alpha} e^{-\Delta V_{ds}/V_{inj}} \\
    & I_{tun}=I_{tun0} e^{(V_{tun}-V_{fg})/V_{ox}} \\
\end{aligned}
\end{equation}
where $I_{d}$ is the drain current, $\alpha$ = 1- $U_{T}$/$V_{inj}$ , $V_{ox}$ and $V_{inj}$ are process dependent parameters.

\subsection{D-STDP Model on a Floating Gate Synapse (FG D-STDP)}
\label{sec:fg_D-STDP}


The weight of the synaptic device can be defined as \cite{fg_stdp}:
\begin{equation}
\begin{aligned}
	\label{eq:weight}
    & w =e^{\frac{-\kappa V_{fg}}{U_T}} \\
\end{aligned}
\end{equation}
Hence, equations to predict the change in FG voltage effectively predict the change in weight.
The following assumptions are made in deriving the theoretical equations \cite{Roshan_IJCNN2014}:

\begin{enumerate}
 \item To ensure small change in weight at very large negative and positive values of $\Delta t$, $V_{g\_init}$ has to be high enough so that $V_{tun\_max}$ - $V_{g\_init}$ is small enough for negligible tunneling. Similarly $\Delta V_g$ = $V_{g\_init}$ - $V_{g\_min}$, should be small enough so that $V_{tun\_init}$ - $V_{g\_min}$ is small enough for negligible tunneling.

 \item Strong coupling from gate to floating gate node where as a weak coupling from tunneling node to floating gate node. This is justified since typically gate capacitance, $C_g>>$ tunneling capacitance, $C_{tun}$.

\item The gate voltage waveform falls to its minimum value instantaneously. In other words, $S_1>>S_2$, $S_3$ as shown in Fig. \ref{fig:STDP_timing_dia}(a).
\end{enumerate}

One difference from the earlier case in \cite{Roshan_IJCNN2014} is that now we only have injection for $\Delta t > 0$ and only tunneling for $\Delta t < 0$ due to the localization of tunneling effects.

We can now derive the slow time scale equation \cite{fg_stdp} for change in FG voltage due to tunneling and injection as:
\begin{equation}
\begin{aligned}
	\label{eq:slow_time_eqn}
    & C_{T} \frac {d\overline{V_{fg}}}{dt} =I_{tun} - I_{inj}= C_{T} \frac {d\overline{V_{fg\_tun}}}{dt} - C_{T} \frac {d\overline{V_{fg\_inj}}}{dt}\\
\end{aligned}
\end{equation}
where $C_{T}$ is the total capacitance on the FG node and $\overline{V_{fg}}$ denotes change on a slow time scale.

\subsubsection{Case 1: $\Delta t > 0$}

 First, we consider the case of $\Delta t  > 0 $ i.e the positive axis of STDP curve and combine equations (\ref{eq:FG_phenomenon}) and (\ref{eq:FG_drain_current}) to get:

\begin{equation}
\begin{aligned}
	\label{eq:slow_time_inj_eqn_1}
    & C_{T} \frac {d\overline{V_{fg\_inj}}}{dt} = - I_{inj0} (e^{\kappa (V_{dd}-V_{fg})/U_{T}})^{\alpha} e^{-\Delta V_{ds}/V_{inj}} \\
\end{aligned}
\end{equation}

where $\overline{V_{fg\_inj}}$ is the slow time scale change in $V_{fg}$ due to injection only. Since change in $V_{fg}$ on the RHS happens due to coupling from the gate voltage, we can write:
\begin{equation}
\begin{aligned}
	\label{eq:coupling}
    & V_{fg} = V_{fg\_min} + \frac {C_g} {C_T} S_2 t \\
\end{aligned}
\end{equation}
where $S_2$ is positive slope of $V_g$ and $C_g$ is the capacitance connected between the gate terminal and the floating gate terminal. Here $V_{fg\_min}$ is not same as $V_{g\_min}$ due to initial charge stored on the FG. Substituting equation (\ref{eq:coupling}) in equation (\ref{eq:slow_time_inj_eqn_1}),we get
\begin{equation}
\begin{aligned}
	\label{eq:slow_time_inj_eqn_2}
    & C_{T} \frac {d\overline{V_{fg\_inj}}}{dt} = - A e^{-X t} \\
\end{aligned}
\end{equation}
where
\begin{equation}
\begin{aligned}
	\label{eq:A&X}
    & A = I_{inj0} ( e^{\alpha \kappa (V_{dd}- V_{fg\_min})/ U_{T}} ) ( e^{-\Delta V_{ds}/V_{inj}} ) \\
    & X = \frac {\alpha \kappa C_g S_2}{C_T U_T} \\
\end{aligned}
\end{equation}

Referring to Fig. \ref{fig:STDP_timing_dia}(a), significant amount of injection happens in the time from $\Delta t$ to $\Delta t$ + $T_{d}$ ( circled in red ), where $\Delta V_{ds}$ is constant and significant. Also, since $T_d$ is very small compared to $T_g$, we can assume that $V_g$ is constant during the drain pulse. Hence, we finally get:

\begin{equation}
\begin{aligned}
	\label{eq:slow_time_inj_eqn_4}
    & C_{T} \Delta \overline{V_{fg\_inj}} =- A T_d e^{- X \Delta t} \\
\end{aligned}
\end{equation}
For more details of the derivation, we refer the interested readers to \cite{Roshan_IJCNN2014}.

With reference to Fig. \ref{fig:STDP_timing_dia}(b), since we have modified the tunneling voltage waveform from $V_{tun}$ to $V_{tun\_eff}$ the effect of tunneling is not present in $\Delta t > 0$ . Hence, we have $ \Delta \overline{V_{fg}} = \Delta \overline{V_{fg\_inj}}$.

\subsubsection{Case 2: $\Delta t < 0$}

 Now we consider the case of $\Delta t = (t_{post} - t_{pre}) < 0 $ i.e the negative axis of STDP curve. Similar to what we have done above, let us see the effect of tunneling and injection separately.

For the contribution of injection to $\overline{V_{fg}}$, we could see that injection happens only during the initial small period, $T_d$ of time axis where the drain current of the MOS is almost zero. So we can completely neglect the effect of injection on $\overline{V_{fg}}$ in this case.

Now let us consider the contribution of tunneling to $\overline{V_{fg}}$. Similar to the analysis above, using assumption 1, we have:
\begin{equation}
\begin{aligned}
	\label{eq:slow_time_eqn_tun_neg1}
    & C_{T} \int_0^{\Delta \overline{V_{fg\_tun}}}dV_{fg} = \int_{-\Delta t}^{T_{tun}}I_{tun} dt \\
\end{aligned}
\end{equation}

Also, similar to the case of positive $\Delta t$, here we have:
\begin{equation}
\begin{aligned}
	\label{eq:coupling_1&Vtun_neg_dt}
    & V_{fg} = V_{fg\_min} + \frac {C_g} {C_T} S_2 (t - (-\Delta t)) \\
    & V_{tun} = V_{tun\_max} + S_3 (t-T_{tun\_delay}) \\
    & V_{tun\_eff} =
\begin{cases}
     V_{tun};\quad  -\Delta t < t < - \Delta t + T_{tun\_pulse} \\
     V_{tun\_init}; \quad for\quad other\quad values\quad of\quad t \\
\end{cases}
\end{aligned}
\end{equation}
Substituting equations (\ref{eq:coupling_1&Vtun_neg_dt}) and (\ref{eq:FG_phenomenon}) into equation (\ref{eq:slow_time_eqn_tun_neg1}), we get

\begin{equation}
\begin{aligned}
	\label{eq:slow_time_eqn_tun_neg3}
    C_{T} \Delta \overline{V_{fg\_tun}} &= B e^{\frac {-C_g S_2 \Delta t} {C_T V_{ox}}}\int_{-\Delta t}^{-\Delta t + T_{tun\_pulse}}{e^{Yt}}dt \\
                        &= B' (e^{Y (T_{tun\_pulse} - \Delta t)} - e^{-Y \Delta t}) e^{\frac {-C_g S_2 \Delta t} {C_T V_{ox}}} \\
\end{aligned}
\end{equation}
where
\begin{equation}
\begin{aligned}
	\label{eq:B&Y}
      & Y = \frac {S_3 - \frac {C_g S_2} {C_T}} {V_{ox}}  \\
      & B = I_{tun0} ( e^\frac {V_{tun\_max} - S_3 T_{tun\_delay} - V_{fg\_min}} {V_{ox}} ) \\
      & B'= B/Y \\
\end{aligned}
\end{equation}
Since injection is negligible for $\Delta t < $0, we have $ \Delta \overline{V_{fg}} = \Delta \overline{V_{fg\_tun}}$.

\subsection{T-STDP Model on a Floating Gate Synapse (FG T-STDP)}
\label{sec:fg_T-STDP}


Comparing equation (\ref{eq:doublet_stdp}) with equation (\ref{eq:triplet_stdp}), the extra term in T-STDP compared to the D-STDP model is $A_3^+ e^{(\frac{-\Delta t_2}{\tau_y})} e^{(\frac{-\Delta t_1}{\tau_+})}$, which occurs at the arrival of post-synaptic spike. Intuitively, for the case of triplet of spikes, we can see that there should be some modification to the drain voltage pulse compared to the doublet case since it is related to potentiation and is generated at the post-synaptic spike. There are two ways in which we can have more potentiation due to injection: ($1$) increase $\Delta V_{ds}$ or ($2$) increase injection pulse width $T_d$. Based on this, we propose the following two drain voltage waveform that accounts for triplet spike interactions:
\begin{enumerate}
	\item  Single-pulsed $V_d$: This case implements the entire potentiation term in equation (\ref{eq:triplet_stdp_hippo}) by a single pulse of width $T_d$ at $t_{post}$ (Fig. \ref{fig:STDP_timing_dia}(c) -- however, the voltage $V_d$ depends on the time difference $\Delta t_2$ between successive post spikes. So from Fig. \ref{fig:STDP_timing_dia}(c) expression for $V_d$ becomes:
	
	\begin{equation}
	\begin{aligned}
	\label{eq:Vd_1}
	& V_{d} =
	\begin{cases}
	V_{d\_min}-\Delta V_d^s(\Delta t_2);\\ \quad \quad \quad t_{post}(n)<t<t_{post}(n)+T_d \\
	V_{d\_init}; \quad for\quad other\quad values\quad of\quad t \\
	\end{cases}
	\end{aligned}
	\end{equation}
	Note that to satisfy the results of the doublet case, $\Delta V_d^s(\Delta t_2) \to $ 0; as $\Delta t_2 \to \infty $.
	
	\item  Double-pulsed $V_d$: The double-pulsed $V_d$ waveform (Fig. \ref{fig:STDP_timing_dia}(c)) implements the doublet term with the first pulse of width $T_d$ and amplitude same as in FG D-STDP. It then creates the extra triplet term with the following pulse again of width $T_d$ but with different amplitude. The entire waveform is given by:
	
	\begin{equation}
	\begin{aligned}
	\label{eq:Vd_2}
	& V_{d} =
	\begin{cases}
	V_{d\_min};  \quad  t_{post}(n) < t < t_{post}(n)+T_d \\
	V_{d\_min}-\Delta V_d^d(\Delta t_2); \\ \quad \quad \quad t_{post}(n)+T_d<t<t_{post}(n)+2*T_d \\
	V_{d\_init};  \quad for\quad other\quad values\quad of\quad t \\
	\end{cases}
	\end{aligned}
	\end{equation}
	Again, note that to satisfy the results of the doublet case, $V_{d\_min}$-$\Delta V_d^d(\Delta t_2) \to $ $V_{d\_init}$; as $\Delta t_2 \to \infty $ i.e second pulse in double-pulsed $V_d$ vanishes as $\Delta t_2 \to \infty $.
\end{enumerate}

As can be seen above, both cases of drain voltage waveform will require some different method to calculate $\Delta V^s_d(\Delta t_2)$ and $\Delta V^d_d(\Delta t_2)$ to match results from the T-STDP model with desired parameters. So, next we do a comparison of the FG phenomenon and T-STDP model to understand the kind of modification to be done. Here, we consider only the FG phenomenon of injection that happens at post-synaptic pulses, since our modification has localized tunneling to pre-synaptic events. Then, we can equate this to the weight change from T-STDP model for any specific case (e.g. hippocampal culture data set in \cite{pfister_triplet}) to extract the exact parameters of the pulse drain waveform. Here, we will express the weight change for T-STDP normalized to D-STDP for ease of understanding; this is intuitively good since T-STDP has modifications on top of D-STDP. Let us consider this case by case.

\begin{enumerate}
\item Case $1$: FG T-STDP for spike doublet inputs

The FG phenomenon on doublets of spike is exactly similar to the derivation shown in previous subsection \ref{sec:fg_D-STDP} for single and double pulsed $V_d$ cases. Therefore,

\begin{equation}
\begin{aligned}
	\label{eq:Vfg_inj_doublet}
    & \Delta \overline{V_{fg\_inj}^{doublet}} = A e^{-X \Delta t} e^{(V_{dd}- V_{d\_min})/V_{inj}} \\
\end{aligned}
\end{equation}
where
\begin{equation}
\begin{aligned}
	\label{eq:A&X}
    & A = \frac {I_{inj0} (e^{\alpha \kappa (V_{dd}- V_{fg\_min})/ U_{T}}) (e^{-X T_d}-1)} {C_T X} \\
    & X = \frac {\alpha \kappa C_g S_2}{C_T U_T} \\
\end{aligned}
\end{equation}

\item Case $2$: FG T-STDP for spike triplet inputs

\begin{enumerate}
\item Single-pulsed $V_d$: Similar to the mathematics in case $1$, we get:
\begin{equation}
\begin{aligned}
	\label{eq:Vfg_inj_triplet_1}
    & \Delta \overline{V_{fg\_inj}^{triplet}} = A e^{-X \Delta t} e^\frac{\Delta V_d^s(\Delta t_2)}{V_{inj}} e^\frac{(V_{dd}- V_{d\_min})}{V_{inj}} \\
\end{aligned}
\end{equation}
where A and X are same as above.

The ratio of change in weight due to FG T-STDP on triplets and doublets of spike is given as,
\begin{equation}
\begin{aligned}
	\label{eq:FG_ratio_1}
	   & Y_{fg}=\frac {\Delta W_{inj}^{triplet}}{\Delta W_{inj}^{doublet}} =\frac {\Delta \overline{V_{fg\_{inj}}^{triplet}}}{\Delta \overline{V_{fg\_{inj}}^{doublet}}} =e^\frac{\Delta V_d^s(\Delta t_2)}{V_{inj}}\\
\end{aligned}
\end{equation}

\item Double-pulsed $V_d$: Again, similar to the mathematics in case $1$, we get:
\begin{equation}
\begin{aligned}
	\label{eq:Vfg_inj_triplet_2}
    & \Delta \overline{V_{fg\_inj}^{triplet}} = \Delta \overline{V_{fg\_inj}^{doublet}}(1+e^{-X T_d}e^\frac{\Delta V_d^d(\Delta t_2)}{V_{inj}}) \\
\end{aligned}
\end{equation}
where X is same as above. Here, the term $e^{-X T_d}$ arises since the gate voltage has changed a little bit at the start of the second pulse compared to the first. Since $e^{-X T_d} \approx 1$ due to small value of $T_d$, the ratio of change in weight due to FG T-STDP on triplets and doublets of spike can be written as:
\begin{equation}
\begin{aligned}
	\label{eq:FG_ratio_2}
	   & Y_{fg}=\frac {\Delta W_{inj}^{triplet}}{\Delta W_{inj}^{doublet}} =\frac {\Delta \overline{V_{fg\_{inj}}^{triplet}}}{\Delta \overline{V_{fg\_{inj}}^{doublet}}} = 1 + e^\frac{\Delta V_d^d(\Delta t_2)}{V_{inj}}\\
\end{aligned}
\end{equation}

\end{enumerate}

\item T-STDP rule for spike doublets and triplets

For the case of T-STDP and D-STDP models the ratio Y can be obtained directly from equation (\ref{eq:triplet_stdp_hippo}) as:
\begin{equation}
\begin{aligned}
	\label{eq:STDP_ratio}
	   & Y_{theory}=\frac {\Delta W^{+}_{triplet}}{\Delta W^{+}_{doublet}} =1+ \frac{A_3^+ e^{(\frac{-\Delta t_2}{\tau_y})}}{A_2^+} \\
\end{aligned}
\end{equation}
Now, we can equate the $Y_{fg}$ values in equations (\ref{eq:FG_ratio_1}) and (\ref{eq:FG_ratio_2}) with the $Y_{theory}$ value in equation (\ref{eq:STDP_ratio}) to get the desired drain voltage parameters.

\begin{enumerate}
\item Single-pulsed $V_d$:

\begin{equation}
\begin{aligned}
	\label{eq:FG_STDP_ratio_1}
	    & Y_{fg} =Y_{theory}\\
	   &\implies e^{\Delta V_d^s(\Delta t_2)/V_{inj}} =1+ \frac{A_3^+ e^{(\frac{-\Delta t_2}{\tau_y})}}{A_2^+} \\
	   &\implies \Delta V_d^s(\Delta t_2) = V_{inj} \ln (1+ \frac{A_3^+ e^{(\frac{-\Delta t_2}{\tau_y})}}{A_2^+})
\end{aligned}
\end{equation}

\item Double-pulsed $V_d$:

\begin{equation}
\begin{aligned}
	\label{eq:FG_STDP_ratio_2}
	    & Y_{fg} =Y_{STDP}\\
	   &\implies e^{\Delta V_d^d(\Delta t_2)/V_{inj}} = \frac{A_3^+ e^{(\frac{-\Delta t_2}{\tau_y})}}{A_2^+} \\
	   &\implies \Delta V_d^d(\Delta t_2) = V_{inj} \ln (\frac{A_3^+ e^{(\frac{-\Delta t_2}{\tau_y})}}{A_2^+})
\end{aligned}
\end{equation}

\end{enumerate}
\end{enumerate}

\subsection{Parameter Translation from Theory to FG model}

\begin{figure} \centering			
\includegraphics[scale=0.8]{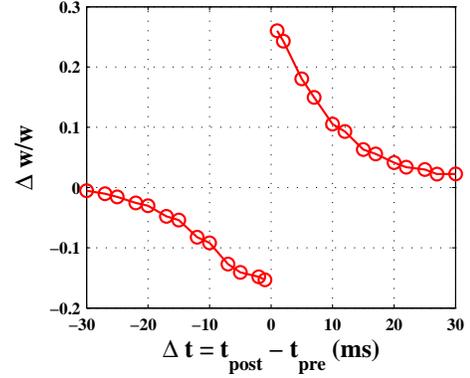} \\
\caption{FG D-STDP learning window: The curve shows learning window obtained from FG synapse using T-STDP rule. Here, $A^+_{fg} \approx 0.26$, $A^-_{fg} \approx -0.15$, $\tau^+_{fg} \approx 25$ms and $\tau^-_{fg} \approx 22$ms }
\label{fig:T-STDP_learn_window}
\end{figure}

Given values of parameters such as $A_2^+$ or $A_3^+$ of the theoretical model, it should be possible to use the above equations (\ref{eq:FG_STDP_ratio_1}) and (\ref{eq:FG_STDP_ratio_2}) to find the values of $\Delta V^s_d$ and $\Delta V^d_d$. However, in our case, the measurement setup restricted the temporal width of voltage waveforms to a maximum value of $300$ms (note the maximum time duration used for $T_{tun}$). Hence, compared to the learning window $\tau^+$ (refer to Fig. \ref{fig:STDP_pfister}) shown in \cite{pfister_triplet}, our learning window $\tau_{fg}^+$ (Fig. \ref{fig:T-STDP_learn_window}) has lesser width. This implies that we need to apply a compression factor on the time scale to match our results with those in \cite{pfister_triplet}. We define the compression factor $r$ as follows:
\begin{equation}
\begin{aligned}
	\label{eq:r}
	   r = \frac {\tau^+}{\tau_{fg}^+}
\end{aligned}
\end{equation}
For the present case, we get the value of $r\approx 2$. Thus, to replicate the weight changes in \cite{pfister_triplet}, we need to reduce all the temporal dimensions in our case by the factor of $r$. This can be formally written as:
\begin{equation}
\begin{aligned}
\label{eq:compression}
\Delta V_d^s(\Delta t_2/r) = V_{inj} \ln (1+ \frac{A_3^+ e^{(\frac{-\Delta t_2}{\tau_y})}}{A_2^+})\\
\Delta V_d^d(\Delta t_2/r) = V_{inj} \ln (\frac{A_3^+ e^{(\frac{-\Delta t_2}{\tau_y})}}{A_2^+})
\end{aligned}
\end{equation}
Next, we show this translation explicitly for the three experimental protocols considered in \cite{pfister_triplet}.
\begin{enumerate}
\item Triplet experiments:

For the different parameters of hippocampal culture data set from table 4 in \cite{pfister_triplet} i.e $A_3^+$ = $9.1$x$10^{-3}$, $A_2^+$ = $4.6$x$10^{-3}$ and $\tau_y = 48$ms, we can obtain from equation (\ref{eq:FG_STDP_ratio_1}) for the case of single-pulsed $V_d$ waveform $\Delta V_d^s(\Delta t_2) \approx 0.24$V for the case of $\Delta t_2 = 10$ms and $\Delta V_d^s(\Delta t_2)\approx 0.21$V for $\Delta t_2 = 20$ms. Similarly, for the case of double-pulsed $V_d$ waveform, we can obtain from equation (\ref{eq:FG_STDP_ratio_2}); $\Delta V_d^d(\Delta t_2)\approx 0.12$V for the case of $\Delta t_2 = 10$ms and $\Delta V_d^d(\Delta t_2) = 66$mV for $\Delta t_2 =20$ms.

The definitions of ($\Delta t_1$ and $\Delta t_2)$ here are similar to \cite{mostafa_triple_2013} and are different compared to the notations given in \cite{pfister_triplet}. For clarity, the difference is shown in Fig. \ref{fig:T-STDP_timing_dia}. As an example, the data points chosen in protocol $2$ are given by $(\Delta t_1 /r,\Delta t_2 /r )$ where $(\Delta t_1,\Delta t_2)$ $\Rightarrow$ (-5,5), (-10,10), (-5,15) and (-15,5) ms corresponds respectively to data points in \cite{pfister_triplet}.

\item Frequency effects of pairing protocol:

In \cite{pfister_triplet}, $\Delta t$ used for the case of frequency effects is $10$ms which is halved to $5$ms for obtaining the measurement results in this paper. Also, $\rho$ Hz mentioned in frequency effects of pairing protocol results of \cite{pfister_triplet} will be equal to $r\times\rho$ Hz in our measurement results.

\item Quadruplet experiments:

Quadruplet experiments contain two temporal variables $\Delta t$ and $T$ as shown in Fig. \ref{fig:STDP_timing_dia}(c). $\Delta t = 5$ms in \cite{pfister_triplet}--hence, we have used $\Delta t = 2.5$ms for measurements. Then the chip measurement values for quadruplets are taken for different values of $T/r$ up to $80$ms. For comparing these measurement results with mathematical model, we simulated the mathematical model for double the value of $T$ i.e up to $160$ms.

\end{enumerate}

\section{Results}
\label{sec:result}

\begin{figure*}
	\begin{minipage}[b]{0.5\textwidth}
		\centering
		\includegraphics[width=6cm, height=4cm]{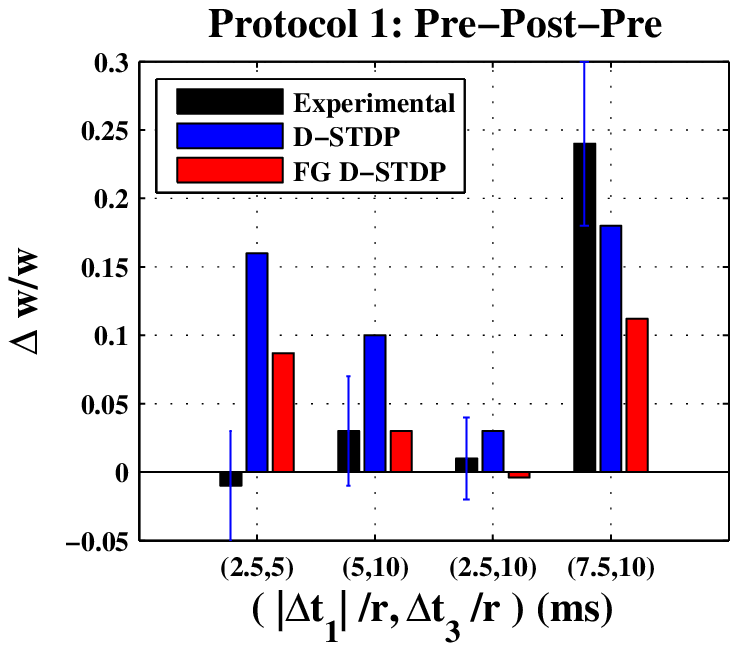} \\ (a)
	\end{minipage}
	\begin{minipage}[b]{0.5\textwidth}
		\centering
		\includegraphics[width=6cm, height=4cm]{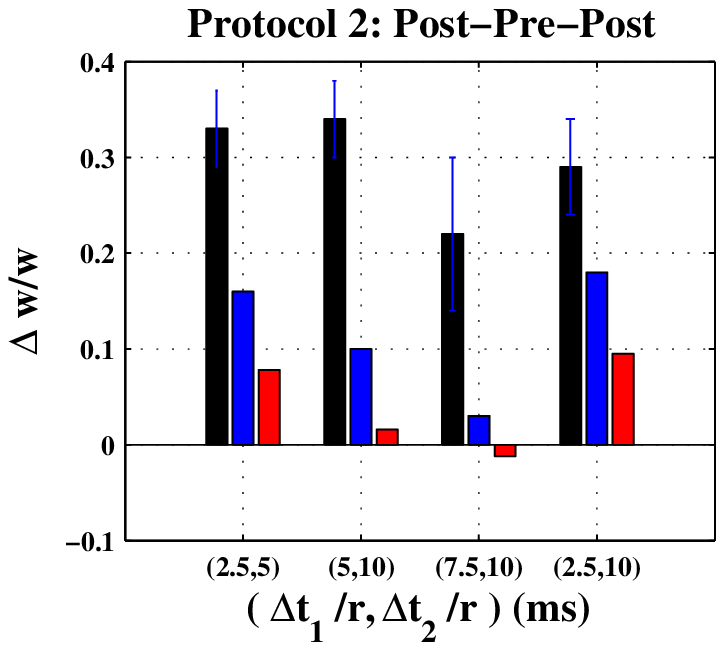} \\ (b)
	\end{minipage}
	\begin{minipage}[b]{0.5\textwidth}
		\centering
		\includegraphics[width=6cm, height=4cm]{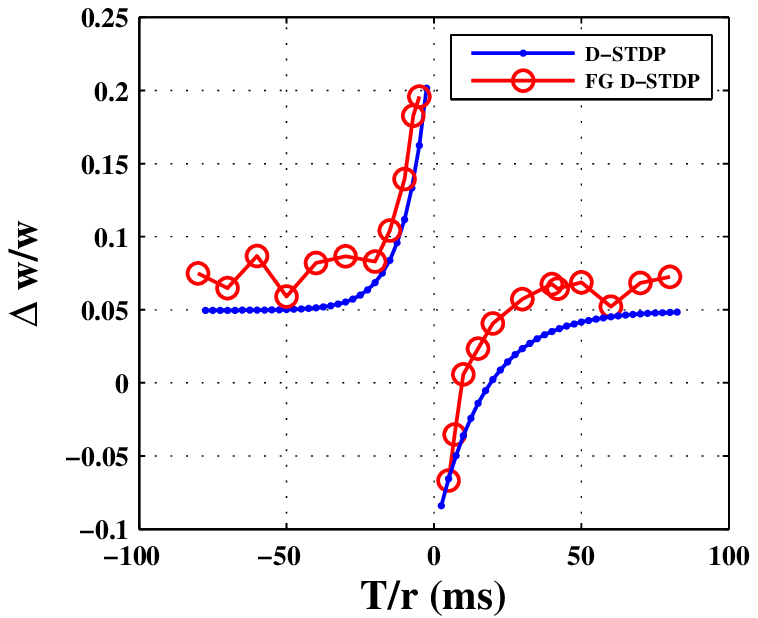} \\ (c)
	\end{minipage}
	\begin{minipage}[b]{0.5\textwidth}
		\centering
		\includegraphics[width=6cm, height=4cm]{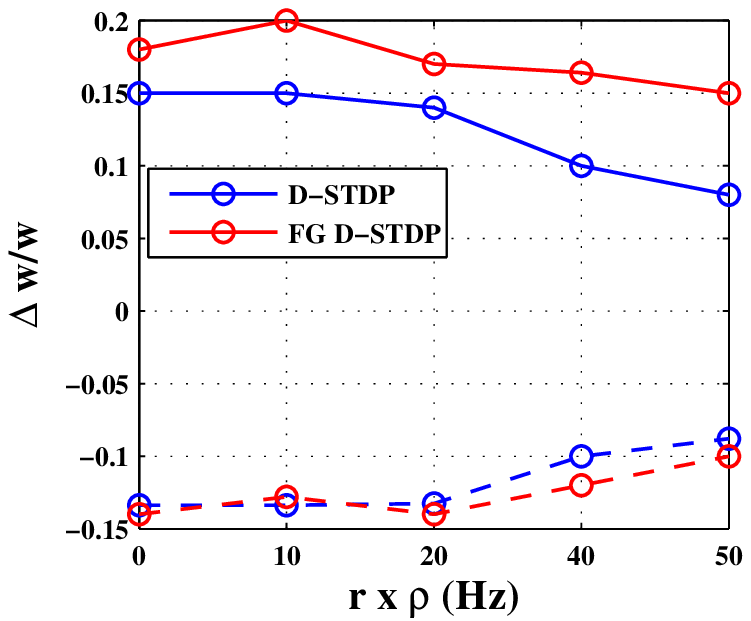} \\ (d)
	\end{minipage}
	\caption{Failure of FG D-STDP: In all four subgraphs, black bars denote experimental data (taken from \cite{pfister_triplet}), blue bars and lines, mentioned as D-STDP correspond to mathematical model proposed in \cite{pfister_triplet} and red lines correspond to measurement results. (a) The bar graphs shows the comparison of results obtained from FG D-STDP (red) with the biological experimental results (black) and mathematical model (blue) for two cases of triplets, case $1$: Pre-Post-Pre ( $|\Delta t_1| /r$, $\Delta t_3 /r$) and (b) case $2$: Post-Pre-Post ( $\Delta t_1 /r$, $\Delta t_2 /r$). FG D-STDP model (red bars) cannot replicate the experimental results shown in black bars. (c) FG D-STDP fails to reproduce quadruplet experiments. (d) Weight change in FG D-STDP rule as a function of frequency, $r \times \rho$. Note that, thick line correspond to $\Delta t = 5$ms and dash line correspond to $\Delta t = -5$ms.} \label{fig:D-STDP_fail}
\end{figure*}
The measurement results shown in this section is obtained from a single FG synapse fabricated in TSMC 0.35$\mu$m CMOS process with $C_g=5.5$ pF. Though not optimized for synaptic density, the results here do serve as a proof of concept for our FG T-STDP implementation.

\subsection{FG D-STDP learning window}
\label{sec:result_learn_window}
The first set of testing on FG synapse is to check whether the FG T-STDP can reproduce D-STDP learning window. Fig. \ref{fig:T-STDP_learn_window} shows the D-STDP learning window obtained from FG synapse. Thus for obtaining the result, we set $V_{d\_min}$ as 0.3V (here, $\Delta V_d (\Delta t_2)$ = 0) w.r.t the mathematical analysis done in the section \ref{sec:fg_T-STDP} and other voltage specifications are as given in Fig. \ref{fig:STDP_timing_dia}(b) i.e $V_{dd}$ = 5.2V, $V_{g\_init}$ = 3.3V, $V_{g\_min}$ = 2.5V, $T_g$ = 100ms, $V_{tun\_init}$ = 5.4V, $V_{tun\_max}$ = 16.5V, $T_{tun}$ = 300ms, $T_{tun\_delay}$ = 1ms, $T_{tun\_pulse}$ = 2ms, $V_{d\_init}$ = 5V and $T_d$ = 500us.

\subsection{Failure of FG D-STDP}
\label{sec:result_fail}
We applied D-STDP rule on FG synapse to the pre- and post-synaptic spike pairs as done in \cite{pfister_triplet}. For obtaining the measurement results in Fig. \ref{fig:D-STDP_fail}, we set the voltage and timing parameters as given in Fig. \ref{fig:STDP_timing_dia}(b) i.e, same as mentioned in subsection \ref{sec:result_learn_window}.

\subsubsection{FG D-STDP fail to reproduce frequency effects}

For obtaining measurement results, we have set $\Delta t = \pm5$ms as shown in Fig. \ref{fig:STDP_timing_dia}(c). The measurement results are plotted as weight change in D-STDP rule as a function of frequency, $\rho$. The measurement results (red lines) shown in Fig. \ref{fig:D-STDP_fail}(d), almost follows the mathematical D-STDP model (blue lines) in \cite{pfister_triplet}. As mentioned in \cite{pfister_triplet}, this can be attributed to the following reasons. First, As pointed out in \cite{sjostrom}, at low repetition frequency, $\rho$, there is no potentiation. This cannot be captured by FG D-STDP, because pulsed drain voltage waveform, $V_d$ at the occurrence of a post-synaptic spike after a pre-synaptic spike by few milliseconds induces LTP. Second, In experiments, for $\Delta t>$0, potentiation increases when frequency increases. This trend can also not be reproduced by FG D-STDP. In D-STDP model, as soon as the frequency increases, the pre-post spike pairs approach each other and the post-synaptic spike of the first spike pair interact with the pre-synaptic spike of the subsequent spike pair. With increase in frequency, the post-pre spike interaction increases and therefore depress the synapse, which is not observed in experiments.

\subsubsection{FG D-STDP fail to reproduce triplet experiments}

In triplet experiments, as shown in Fig. \ref{fig:D-STDP_fail}(a) and (b), there is a clear asymmetry between two protocols mentioned (black bars). $60$ repetitions of pre-post-pre triplet yields less weight change, whereas $60$ repetitions of post-pre-post triplet yields a weight change of $\sim$30$\%$ (black bars). However, FG D-STDP shown in red bars, predicts almost same result for both protocols, because the mechanism of potentiation and depression happening due to the generation of different voltage waveforms at pre- or post-synaptic spikes are similar in both protocols. Therefore, triplet results cannot be explained by a sum of pre-post injection mechanism and a post-pre tunneling mechanism.

\subsubsection{FG D-STDP fail to reproduce quadruplet experiments}

The asymmetry present in the quadruplet experiments, as shown in Fig. \ref{fig:D-STDP_fail}(c), also causes some problem for FG D-STDP. A quadruplet consists of a pre-post-post-pre sequence where, T$<$0 or a post-pre-pre-post sequence where, T$>$0 as shown in Fig. \ref{fig:STDP_timing_dia}(c). Here, $|T|$ denotes the interval between the first and last pair of spikes within the quadruplet. Sequence, pre-post-post-pre consists of two pre-post interactions and a post-pre interaction whereas; for the sequence, post-pre-pre-post, the opposite occurs i.e two post-pre interactions and only one pre-post interaction. This clearly leads to an asymmetry which is not seen in experiments \cite{pfister_triplet}.

\subsection{Success of FG T-STDP}
\label{sec:result_success}
FG T-STDP is implemented with the help of a drain voltage waveform generator, explained in next section \ref{sec:drain}. As mentioned in the previous section \ref{sec:fg_T-STDP}, the mathematical analysis provides an intuitive understanding of utilization of $V_d$ pulse waveform to obtain the extra triplet term in the T-STDP rule (equation (\ref{eq:triplet_stdp})). As the extra triplet term can be achieved with the two cases of proposed pulse drain voltage waveform, here, we have shown measurement results for both the cases of $V_d$ waveform. The measurement results obtained for both the case of drain voltage waveform satisfy the behavior seen in experimental results \cite{pfister_triplet}. The timing specifications for frequency effects of pairing protocol, triplets of spike and quadruplets are shown in Fig. \ref{fig:STDP_timing_dia}(c).

\begin{figure*}	
            \begin{minipage}[b]{0.5\textwidth}
        \centering
		\includegraphics[width=6cm, height=4cm]{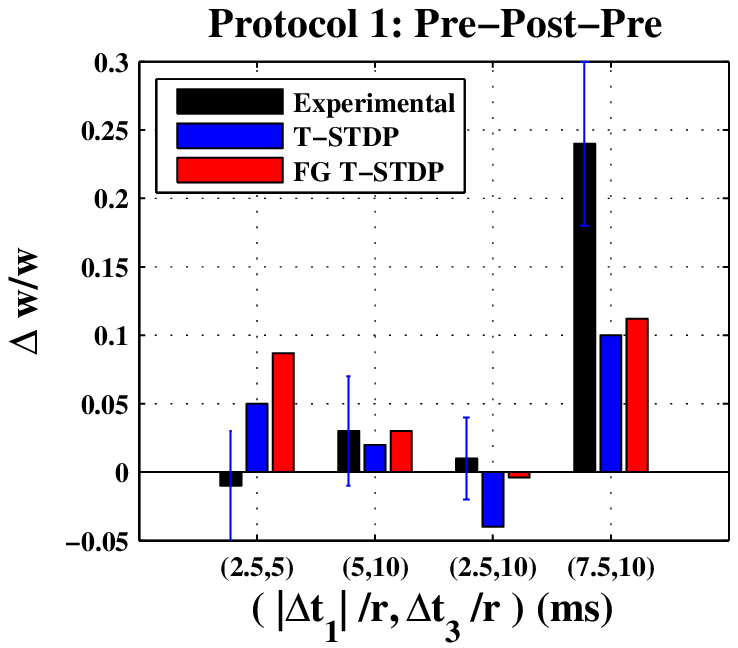} \\ (a)
    \end{minipage}
    \begin{minipage}[b]{0.5\textwidth}
        \centering
		\includegraphics[width=6cm, height=4cm]{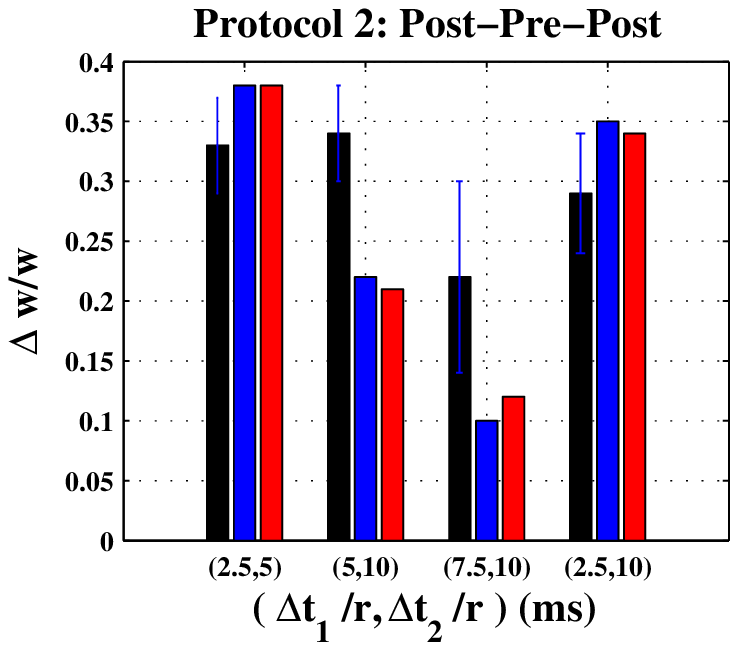} \\ (b)
    \end{minipage}
    \begin{minipage}[b]{0.5\textwidth}
        \centering
		\includegraphics[width=6cm, height=4cm]{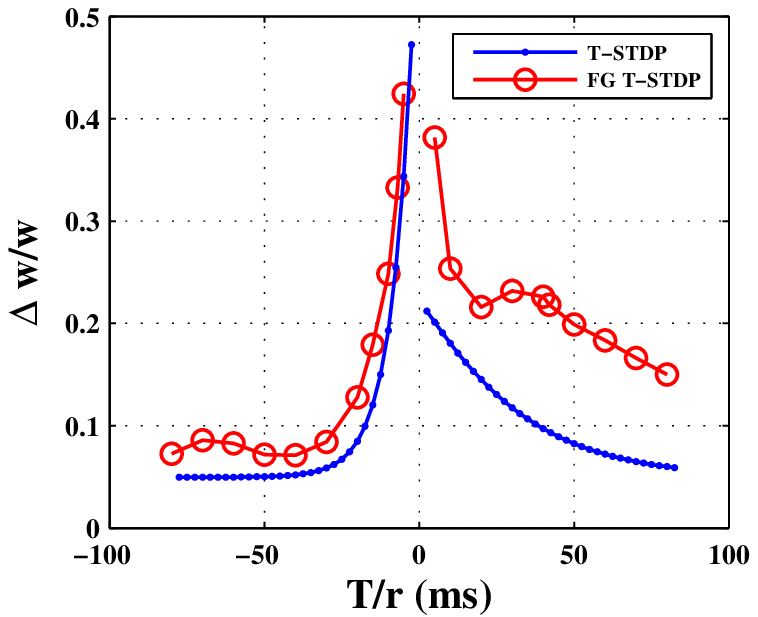} \\ (c)
     \end{minipage}
    \begin{minipage}[b]{0.5\textwidth}
        \centering
		\includegraphics[width=6cm, height=4cm]{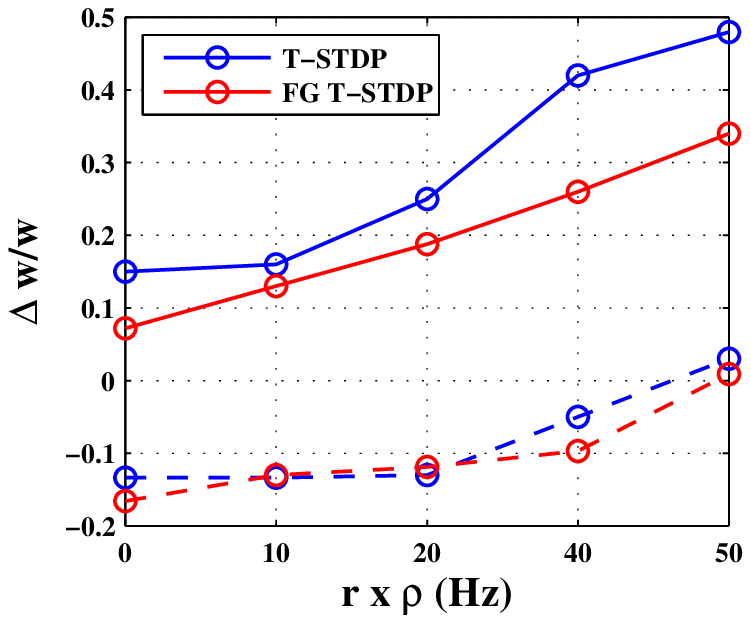} \\ (d)
     \end{minipage}
        \caption{FG T-STDP using single-pulsed drain waveform: In all four subgraphs, black bars denote experimental data (taken from \cite{pfister_triplet}), blue bars and lines, mentioned as T-STDP correspond to Nearest-Spike minimal triplet model proposed in \cite{pfister_triplet} and red lines correspond to measurement results. (a) The bar graphs shows the comparison of results obtained from FG T-STDP (red) with the biological experimental results (black) and mathematical model (blue) for T-STDP protocol $1$:Pre-Post-Pre ( $|\Delta t_1| /r$, $\Delta t_3 /r$). Similar to (a), results for T-STDP protocol $2$:Post-Pre-Post ( $\Delta t_1 /r$, $\Delta t_2 /r$) is shown in (b). FG T-STDP (red bars) can replicate the experimental results shown in black bars. (c) FG T-STDP can also reproduce quadruplet experiments. (d) Weight change in T-STDP rule as a function of frequency, $r \times \rho$. FG T-STDP can reproduce frequency effects compared to FG D-STDP. Note that, thick line correspond to $\Delta t = 5$ms and dash line correspond to $\Delta t = -5$ms.} \label{fig:T_stdp_bar_vd_single}
\end{figure*}

\begin{figure*}	
        \begin{minipage}[b]{0.5\textwidth}
        \centering
		\includegraphics[width=6cm, height=4cm]{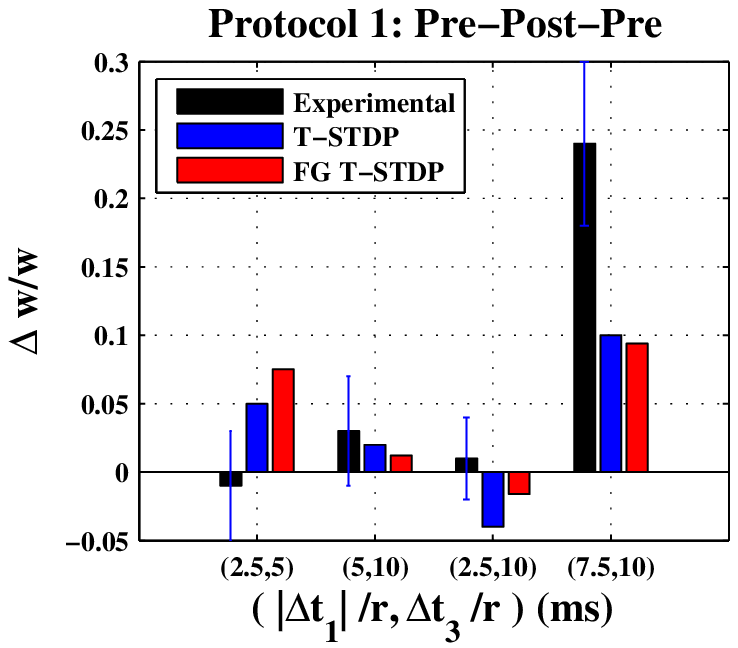} \\ (a)
    \end{minipage}
    \begin{minipage}[b]{0.5\textwidth}
        \centering
		\includegraphics[width=6cm, height=4cm]{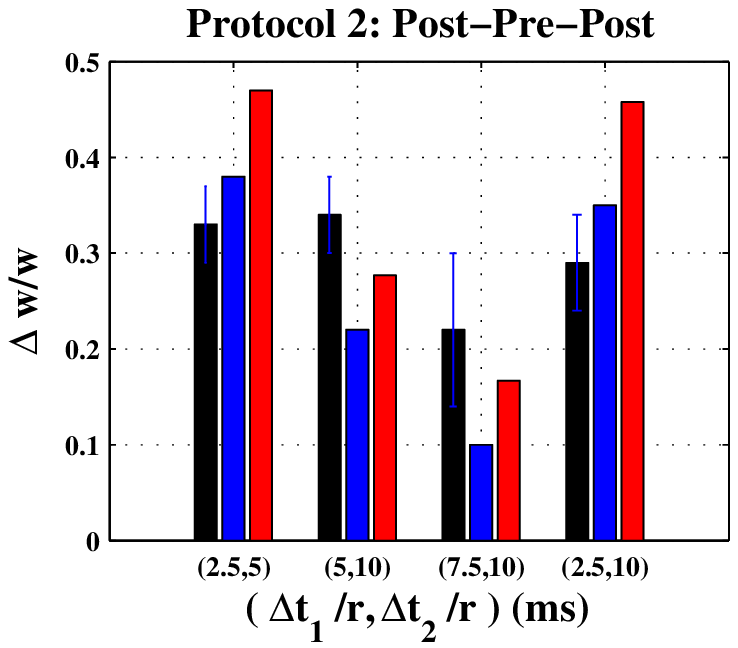} \\ (b)
    \end{minipage}
    \begin{minipage}[b]{0.5\textwidth}
        \centering
		\includegraphics[width=6cm, height=4cm]{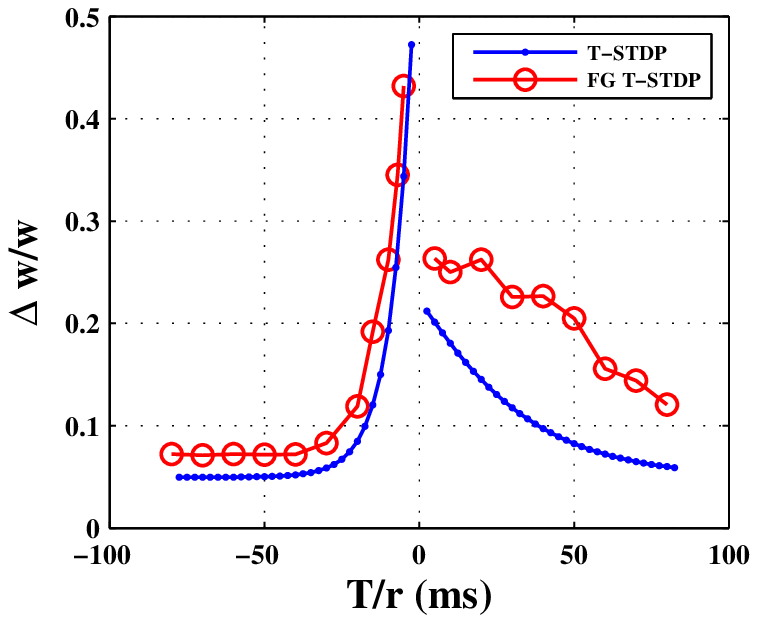} \\ (c)
     \end{minipage}
    \begin{minipage}[b]{0.5\textwidth}
        \centering
        \includegraphics[width=6cm, height=4cm]{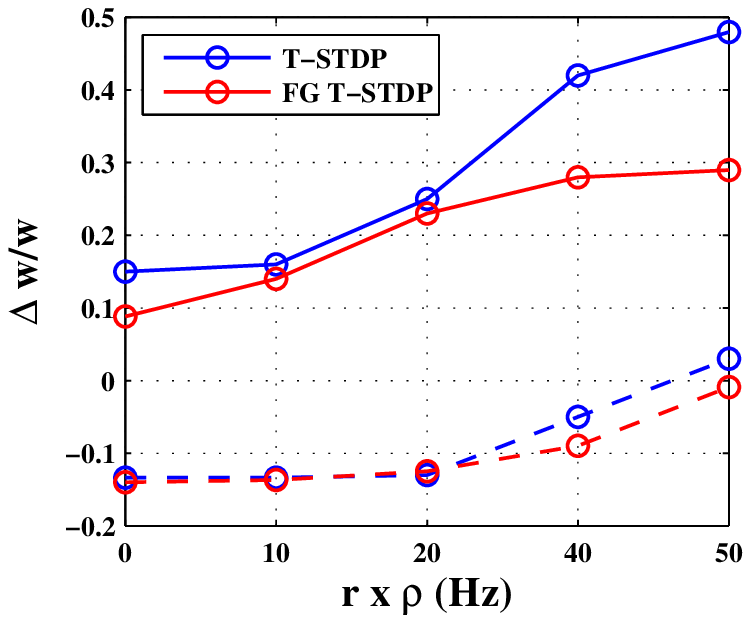} \\ (d)
     \end{minipage}
        \caption{FG T-STDP using double-pulsed drain waveform: In all four subgraphs, black bars denote experimental data (taken from \cite{pfister_triplet}),blue bars and lines, mentioned as T-STDP correspond to Nearest-Spike minimal triplet model proposed in \cite{pfister_triplet} and red lines correspond to measurement results. (a) The bar graphs shows the comparison of results obtained from FG T-STDP (red) with the biological experimental results (black) and mathematical model (blue) for T-STDP protocol $1$:Pre-Post-Pre ( $|\Delta t_1| /r$, $\Delta t_3 /r$). Similar to (a), results for T-STDP protocol $2$:Post-Pre-Post ( $\Delta t_1 /r$, $\Delta t_2 /r$) is shown in (b). FG T-STDP (red bars) can replicate the experimental results shown in black bars. (c) FG T-STDP can also reproduce quadruplet experiments. (d) Weight change in T-STDP rule as a function of frequency, $r \times \rho$. FG T-STDP can reproduce frequency effects compared to FG D-STDP. Note that, thick line correspond to $\Delta t = 5$ms and dash line correspond to $\Delta t = -5$ms.} \label{fig:T_stdp_bar_vd_double}
\end{figure*}

\subsubsection{FG T-STDP can reproduce triplet and quadruplet experiments}

T-STDP model on floating gate synapse does not only reproduce the learning window (Fig. \ref{fig:T-STDP_learn_window}), but also it can reproduce most of the triplet and quadruplet experiments as shown in Fig. \ref{fig:T_stdp_bar_vd_single}(a),(b) and (c) and Fig. \ref{fig:T_stdp_bar_vd_double}(a),(b) and (c). The voltage and timing parameters for the case of single-pulsed drain voltage waveform are given in Fig. \ref{fig:STDP_timing_dia}(b) i.e, same as mentioned in subsection \ref{sec:result_learn_window}. Here, $V_{d\_min}$ = 0.3V is set with respect to $\Delta V_d(\Delta t_2)$ value (equation (\ref{eq:FG_STDP_ratio_1})), for parameters of Nearest-Spike minimal model, hippocampal culture data set, from table 4 in \cite{pfister_triplet} i.e $A_3^+$ = 9.1x$10^{-3}$, $A_2^+$ = 4.6x$10^{-3}$ and $\tau_y$ = 48ms. For the case of double-pulsed drain voltage waveform, voltage and timing parameters are also same as in the case of single pulse but, $\Delta V_d(\Delta t_2)$ is calculated using equation (\ref{eq:FG_STDP_ratio_2}). In the triplet measurement results the red bars can almost capture the experimental black bars i.e more potentiation in the case of protocol $1$ and less potentiation in the case of protocol $2$. The blue bars shows the result of the mathematical model in \cite{pfister_triplet}. The quadruplet measurement results also replicate the symmetry seen in the experimental results \cite{pfister_triplet}.

\subsubsection{FG T-STDP model can reproduce frequency effects}

FG T-STDP shown in red lines in Fig. \ref{fig:T_stdp_bar_vd_single}(d) and Fig. \ref{fig:T_stdp_bar_vd_double}(d) ameliorated the results obtained with FG D-STDP (Fig. \ref{fig:D-STDP_fail}(d)). The trend of increase in potentiation with frequency is seen for both cases of pulse drain voltage waveform: single and double pulse. The voltage and timing parameters for the case of single-pulsed drain voltage waveform are also same as mentioned in subsection \ref{sec:result_learn_window} but, with a difference of $V_{d\_min}$ = 0.5V instead of $0.3$V. $V_{d\_min}$ is set with respect to the $\Delta V_d(\Delta t_2)$ value, (equation (\ref{eq:FG_STDP_ratio_1})), for parameters of $A_3^+$ = 28.7x$10^{-3}$, $A_2^+$ = 4.6x$10^{-3}$ and $\tau_y$ = 48ms. Note that, we have used $A_3^+$ = 28.7x$10^{-3}$, which is almost thrice compared to $A_3^+$ = 9.1x$10^{-3}$ of the hippocampal culture data set, to ensure more potentiation with increase in frequency. Also note that, the measurement results for frequency effect of pairing protocol is obtained with the help of hippocampal culture data set instead of visual cortex data set mentioned in \cite{pfister_triplet}. Again, for the case of double-pulsed drain voltage waveform, voltage and timing parameters are same as in the case of single pulse but, $\Delta V_d(\Delta t_2)$ is calculated using equation (\ref{eq:FG_STDP_ratio_2}), for parameters of $A_3^+$ = 28.7x$10^{-3}$, $A_2^+$ = 4.6x$10^{-3}$ and $\tau_y$ = 48ms. As in \cite{pfister_triplet}, here also we have a limitation i.e the absence of potentiation at low frequency is not observed in the measurement results in the case of single-pulsed and double-pulsed drain waveform (Fig. \ref{fig:T_stdp_bar_vd_single}(d) and Fig. \ref{fig:T_stdp_bar_vd_double}(d)). This is because the single pulse at the first post-synaptic spike itself is enough to generate some injection.

\section{DRAIN VOLTAGE GENERATOR: VLSI IMPLEMENTATION}
\label{sec:drain}

The drain voltage waveform generator is the important block for generating the voltage pulses according to spike timing as shown in Fig. \ref{fig:FG_mux_tun}. The input to the generator is a post-synaptic pulse from a neuron as shown in Fig. \ref{fig:Neu_Syn}. The output pulse from the generator is fed back to the drain terminal of FG synapse. The number of drain waveform generators in a system depend on the number of neurons present in that neural network architecture. We propose circuits for single- and double-pulsed drain voltage waveform generator below along with their SPICE simulation results.

\subsection{VLSI implementation of single-pulsed drain voltage waveform}
\begin{figure}	\centering
	    \centering
			\includegraphics[width=8cm, height=9cm]{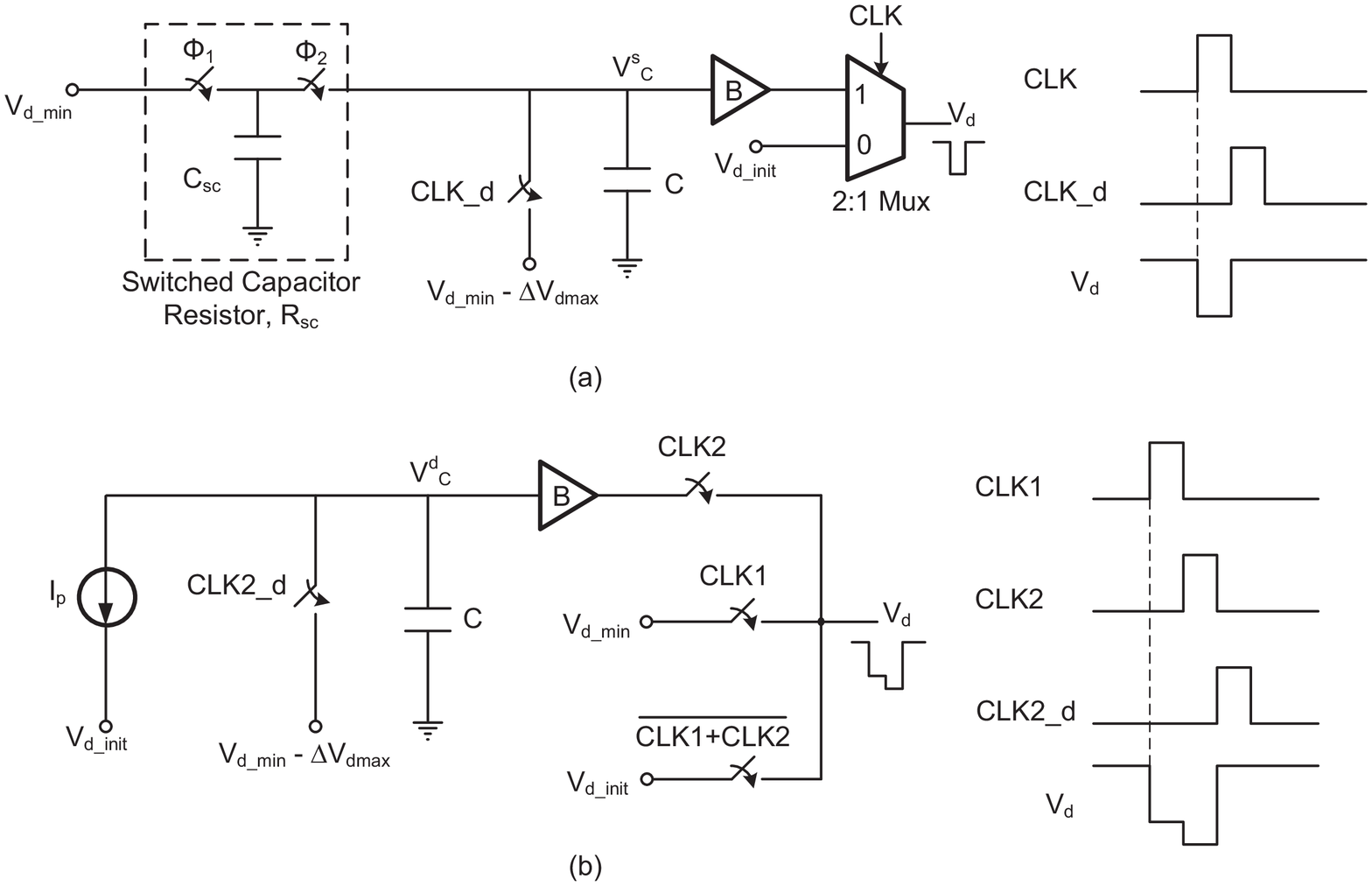} \\
	\caption{VLSI implementations: Circuit implementation of (a) single-pulsed drain voltage waveform and (b) double-pulsed drain voltage waveform.}
	\label{fig:circuit_res}
\end{figure}

From equation (\ref{eq:FG_STDP_ratio_1}), we need to create an exponentially decaying voltage trace for $\Delta V_d^s(\Delta t_2)$ for large values of $\Delta t_2$. Also, $V_{d\_min}$ is the default value of $V_d$ when there is no post synaptic pulse for a long time ($\Delta t_2 \to \infty$). In order to create the exponential voltage trace, we can use a capacitor, C and a switched capacitor resistor, $R_{sc}$, where the capacitor charges from the lowest voltage, $V_{d\_min}$ - $\Delta V_{dmax}$ to $V_{d\_min}$ through the resistor (Fig. \ref{fig:circuit_res}(a)). Here, $\Delta V_{dmax}$ is given by the equation (\ref{eq:FG_STDP_ratio_1}), where $\Delta t_2 \to 0 $. The operation of the circuit is as follows: at every post-synaptic pulse denoted as CLK in Fig. \ref{fig:circuit_res}(a), the voltage across the capacitor $V_c^s$ is sampled as $V_d$ through the multiplexer. At other times, the multiplexer enforces $V_d=V_{d\_init}$. A delayed pulse $CLK\_d$ is also generated after the clock pulse such that it does not overlap with CLK. At this pulse, $V_c^s$ is pulled down to $V_{d\_min}$ - $\Delta V_{dmax}$. After this, $V_c^s$ decays back to $V_d\_min$ by discharging through the resistor $R_{sc}$. The decay constant of this circuit is set by the parameter $\tau_y$ in the equation (\ref{eq:FG_STDP_ratio_1}). Thus, the resistance of switched capacitor and capacitance, C are designed as $R_{sc}$ C = $\tau_y$.

The simulation result of $\Delta V_d^s(\Delta t_2)$ obtained from the VLSI circuit is shown in Fig. \ref{fig:sim_res}(a) with comparison to the ideal theoretical plots. For simulation results, we have used C = 1pF, $V_{d\_min}$ = 0.3V, $V_{d\_init}$ = 5V and $V_{d\_min}$ - $\Delta V_{dmax}$ = 25mV for $A_3^+$ = 9.1x$10^{-3}$ and $A_2^+$ = 4.6x$10^{-3}$. For designing switched capacitor resistor, $R_{sc}$ we have:
\begin{equation}
\begin{aligned}
	\label{eq:SC_single}
	    R_{sc} = & \frac {T_{sc}}{C_{sc}} \\
	   R_{sc} C = & \tau_y \\
        \implies \frac {C}{C_{sc}} = & \frac {\tau_y}{T_{sc}}
\end{aligned}
\end{equation}
Here, $C_{cs}$ is the switched capacitor and $1/T_{sc}$ is the frequency of non overlapping clocks used in switched capacitor. $T_{sc}$ is limited by the desired time resolution for $\Delta t_2$, which is around 2ms. Thus, C/$C_{sc}$ = 24 and $C_{sc} \approx $ 42fF for $\tau_y$ = 48ms. It can be seen that the circuit simulation does not exactly match the theory for moderate values of $\Delta t_2$ due to our approximation of equation (\ref{eq:FG_STDP_ratio_1}) by an exponential.

\subsection{VLSI implementation of double-pulsed drain voltage waveform}

\begin{figure}	\centering
	\includegraphics[width=6cm, height=4cm]{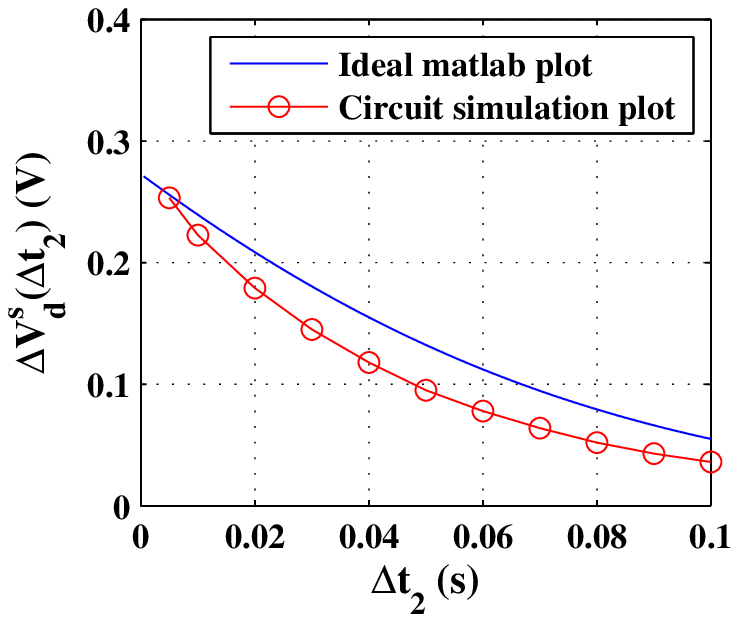} \\ (a) \\
	\centering
	\includegraphics[width=6cm, height=4cm]{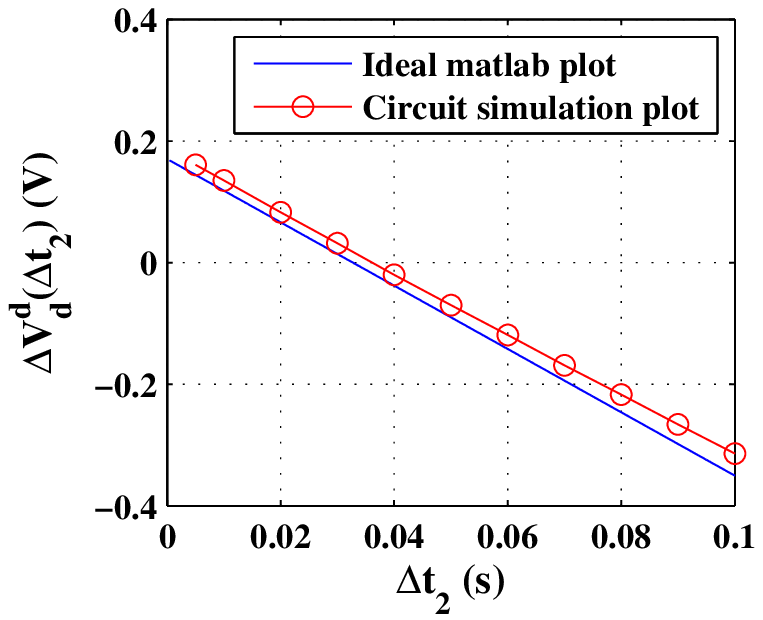} \\ (b)
	\caption{SPICE Simulation results: (a) The simulation result of the single-pulsed drain voltage waveform circuit. The result shows the variation of $\Delta V_d^s(\Delta t_2)$ with respect to $\Delta t_2$. The ideal MatLab simulation result is also included for comparison. (d) The simulation result of the double-pulsed drain voltage waveform circuit.}
	\label{fig:sim_res}
\end{figure}

Fig. \ref{fig:circuit_res}(b) shows the circuit implementation of double-pulsed drain voltage waveform according to equations (\ref{eq:FG_STDP_ratio_2}) and (\ref{eq:Vd_2}). The circuit operation is exactly similar to single pulsed drain waveform generator except that $\Delta V_d^d(\Delta t_2)$ is linearly dependent on $\Delta t_2$. Hence, the resistor is replaced with a current source, $I_p$. Since, we need to create double pulse, an extra clock is used here. CLK$1$ is to create the first pulse, during which $V_d$ = $V_{d\_min}$ and CLK$2$ is to sample the voltage, $V_c^d$ across capacitor, C on to the output node, $V_d$ in order to create the second pulse. Whenever there is no post-synaptic spike, $V_d$ is held at $V_{d\_init}$ through the switch controlled by NOR of CLK$1$ and CLK$2$. For deciding the value of $I_p$; simplifying equation (\ref{eq:FG_STDP_ratio_2}), we get:
\begin{equation}
\begin{aligned}
	\label{eq:IP_double_Vd}
        \Delta V_d^d(\Delta t_2) &= V_{inj} \ln (\frac{A_3^+}{A_2^+}) - V_{inj} \frac{\Delta t_2}{\tau_y} \\
	  \end{aligned}
\end{equation}
also from capacitor charging,
\begin{equation}
\begin{aligned}
	\label{eq:IP_double_C}
	    C \frac{dV}{dt} = & I_p \\
	  \end{aligned}
\end{equation}
Thus equating both the slopes of equations (\ref{eq:IP_double_Vd}) and (\ref{eq:IP_double_C}) , we get:
\begin{equation}
\begin{aligned}
	\label{eq:IP_double}
	    \frac{I_p}{C} = & -\frac{V_{inj}}{\tau_y} \\
	  \end{aligned}
\end{equation}
Hence, $I_p$ = -5.2 pA for $\tau_y$ = 48ms and $V_{inj}$ = 0.25V. The simulation result of $\Delta V_d^d(\Delta t_2)$ obtained from the VLSI circuit is shown in Fig. \ref{fig:sim_res}(b) with comparison to the ideal MATLAB plots. For simulation results, we have used C = 1pF, $V_{d\_min}$ = 0.3V, $V_{d\_init}$ = 5V and $V_{d\_min}$ - $\Delta V_{dmax}$ = 125mV for $A_3^+$ = 9.1x$10^{-3}$ and $A_2^+$ = 4.6x$10^{-3}$. It is easily seen that there is a much better match between the circuit and MATLAB simulation in this case due to the simpler functional form of $\Delta V_d^d$.

\section{Discussion}
\label{discussion}

\begin{figure} \centering			
\includegraphics[width=8cm, height=8cm]{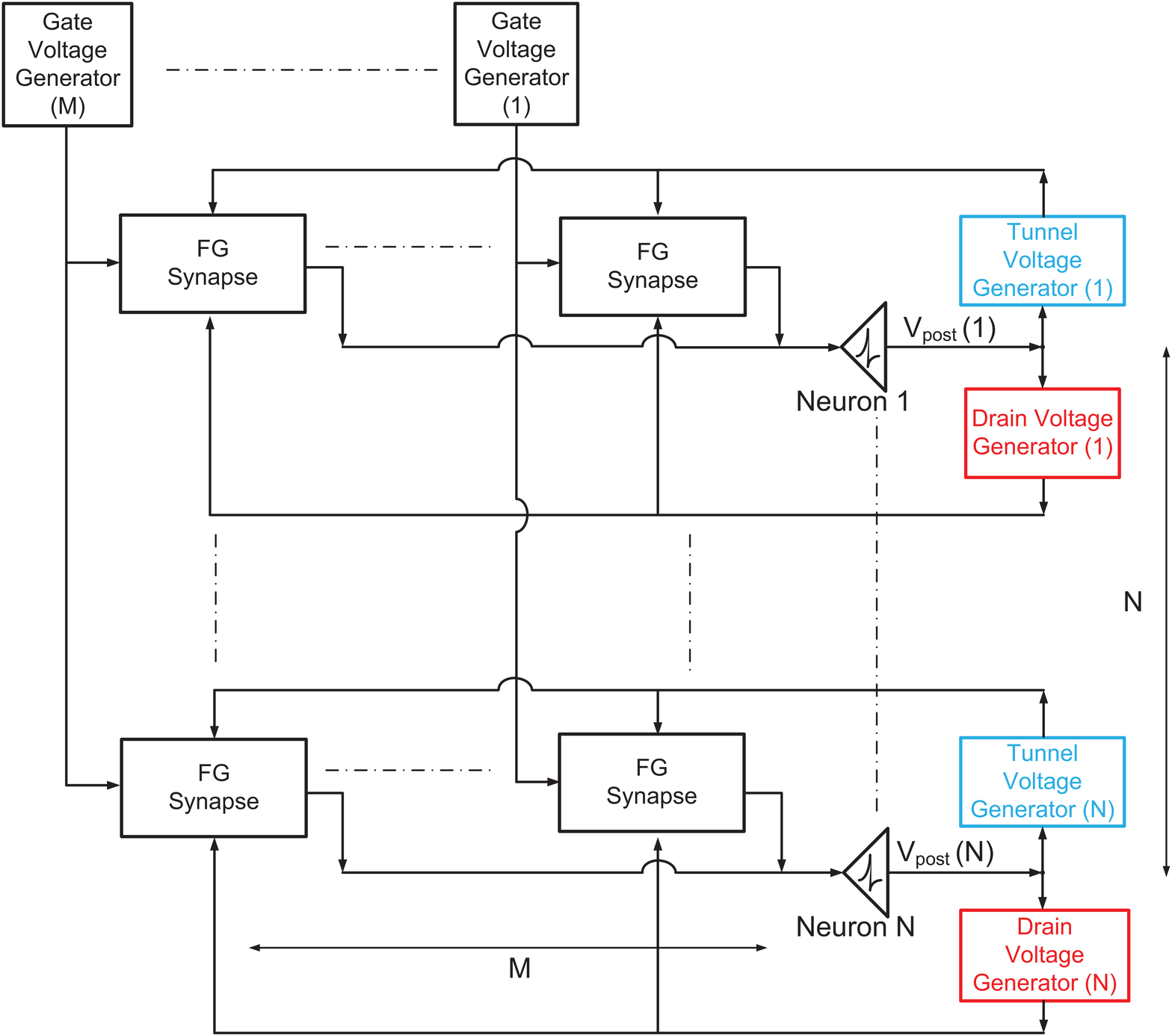}
\caption{System level architecture: The system level implementation of a network with M$\times$N synapses and N neurons.}
\label{fig:Neu_Syn}
\end{figure}

\begin{table*}[ht]
	\centering
	\begin{minipage}[b]{1\textwidth}
		\centering
\caption{Comparison of Learning Synapses in VLSI}\label{synapse_area}
		\begin{tabular}{|c|c|c|c|c|c|c|}
			\hline
			Reference & Process & Synapse & Normalized & Weight storage & Plasticity & Chip  \\
			& technology (nm) & area ($\mu$$m^2$) & Area\footnote{Normalized area is termed as the ratio of the synapse area to the square of the minimum transistor dimension in that process technology.} & (precision) & rule & measurement \\
			\hline
			This paper & 350 & 6337 & 51730 & Floating gate ($>$ 10 bits) & T-STDP & Yes\\
			\cite{mostafa_triple_2013} & 350 & - & - & Capacitor (Analog) & T-STDP & No \\
			\cite{brink_learningfg} & 350 & 133 & 1088 & Floating gate ($>$ 10 bits) & STDP & Yes\\
			\cite{Bamford2012} & 350 & 400 & 3265 & Capacitor (Analog) & STDP & Yes\\
			\cite{fg_stdp} & 350 & 100 & 816 & Floating gate ($>$ 10 bits) & STDP & Yes \\
			\cite{Tanaka2009} &  250 &  5000 & 80000 & Capacitor (Analog) & STDP & Yes \\
			\cite{Koickal_Tara} & 600 & 72000 & 200000 & Capacitor (Analog) & STDP & Yes\\
			\cite{giacomo_spikestdp} &  800 &  4495 & 7023 & Capacitor (bistable) & STDP & Yes \\
			\cite{arthur_gamma} &  250 &  238 & 3810 & SRAM (1 bit) & STDP & Yes \\
			\cite{Schemmel2006} &  180 &  108 & 3338 & SRAM (4 bits) & STDP  & Yes \\
			\cite{Mitra2009} &  350 &  3000 & 24490 & Capacitor (bistable) & SDSP & Yes \\
			\hline
		\end{tabular}
	\end{minipage}
\end{table*}


  An important consideration for synapse designs is scalability to large arrays. Figure \ref{fig:Neu_Syn} shows the system level architecture of a neuromorphic hardware device with $N$ neurons, $M$ inputs and $M$x$N$ synapses (number of synapses outnumbered compared to number of neurons). It shows the connection between FG synapses and neurons and gives an idea about the number of different voltage waveform generators to be used for generating the terminal voltages for the FG synapse shown in Fig. \ref{fig:FG_mux_tun}. Here, for the entire system shown, we need only $N$ drain voltage, $N$  tunnel voltage and $M$ gate voltage waveform generators. Thus the synaptic area overhead compared to earlier implementations \cite{brink_learningfg} is only the added multiplexer for switching tunneling voltages.

  Table \ref{synapse_area} compares this work with other reported implementations of plastic synapses--a detailed review of these circuits can be found in \cite{Mostafa2014}. It can be seen that our work is the first that combines high resolution non-volatile storage with sophisticated plasticity rules. The term normalized area is used to denote the ratio of the synapse area to the square of the process technology. It is a normalized metric to compare the size of a synapse circuit independent of process technology--smaller numbers refer to more compact designs. The floating-gate device used in our test chip is quite large. However,this is not a fundamental problem since we have earlier demonstrated STDP in much smaller floating-gate devices \cite{brink_learningfg}--the only difference of this work with our earlier one is in terms of peripheral circuits to control drain and tunnel waveforms. Hence, after this proof of concept work, we can make a dedicated chip with floating-gates occupying $\approx 100 \mu m^2$ area. Some emerging devices like memristors are also showing promise as a compact learning synapse \cite{WangZQ2012} for spiking systems. Though we are not yet aware of reports of dense arrays of memristive STDP synapses integrated with CMOS neurons in hardware, this seems like a promising area in future when scaling of flash memory or floating gates become limited.

  One of the important aspects of a circuit implementation of a learning rule is the ease with which its parameters can be tuned. We showed in Section \ref{sec:fg_T-STDP} how the parameter $A_3^+$ can be tuned based on $\Delta V_d$. Other learning rule parameters are also directly related to parameters of the control voltage waveforms. For example, $\tau_+$ can be modified using $T_g$, $\tau_-$ using $T_{tun}$ and $A_2^-$ using $T_{tun\_pulse}$ and $V_{tun\_max}$. This is evident from the mathematical analysis of section \ref{sec:fg_basics}. Similarly, from equation (\ref{eq:FG_STDP_ratio_1}) and equation (\ref{eq:FG_STDP_ratio_2}), $A_2^+$, $A_3^+$ and $\tau_y$ can be tuned using $\Delta V_d^s$ or $\Delta V_d^d$, keeping two of them constant at a time. Some results for the variation of learning window to different parameters can be seen in our previous paper \cite{Roshan_TNNLS2015}. Also, other minor changes to the learning rule can be done by modifying the circuits at the neuron and periphery that generate the gate, drain and tunnel waveforms.
	
After this proof of concept, we will continue the work by simulating a spiking neural network (SNN) using SPICE simulations to understand the difference between T-STDP and D-STDP learning rules and then extend our work in future by fabricating chips with thousands of neurons and millions of learning synapses to do tasks like rapid and robust pattern recognition \cite{Huajin_NECO_2013, Huajin_TNNLS_2013}. Such neuromorphic chips that incorporate noise and heterogeneity are useful to understand the principles used by our brain to compute using imprecise elements\cite{giacomo_cognition,neurogrid} as well as for accelerated simulations of neural networks\cite{dudek-2012,spinnaker-jssc}. Moreover, we hope to use neuromorphic systems as the ``brain" for real-time behaving systems like robots\cite{giacomo_cognition} where both low-power dissipation and real-time operation are necessary. In these cases, using a traditional computer for the implementation is inefficient due to the mismatch between Von-Neumann computing model of digital computers and the massively parallel analog computing of the brain where memory and computing are closely intermixed\cite{mead-neuromorphic}. Hence, it is useful to be able to mimic biological neural networks closely in ciruits to enable experimental paradigms as well as low-power intelligence.

\section{Conclusion}
\label{conclusion}
We have presented a spike triplet based learning rule using a single FG transistor as the synapse for VLSI spiking neural networks. The spike triplet affects the setting of drain voltage--we presented a single pulse and a double pulse drain voltage method to obtain the desired dependence of weight on spike timing. We presented a method to calculate the parameters of the drain voltage pulse to obtain results matched to the original theoretical T-STDP rule. We also show FG measurement results in comparison with the biological experimental observations for (1) original doublet protocol, (2) two protocols of spike triplets, (3) frequency effects of pairing protocol and (4) quadruplet experiments. The failure of FG D-STDP rule in replicating the biological results is also included. Possible hardware implementations of drain voltage waveform generator are also proposed and verified through SPICE simulation results. It was shown that the voltage waveform for double pulse case can be generated more accurately due to its simplistic nature.

\section*{Acknowledgement}
\label{acknowlegement}
Financial support from MOE through grant ARC 8/13 is acknowledged. The authors thank Prof. Jennifer Hasler for providing access to FG transistor.

\bibliographystyle{IEEEbib}
\bibliography{ref1}

\begin{thebibliography}{10}

\bibitem{Markram1997}
H.~Markram, J.~L{\''{u}}bke, M.~Frotscher, and B.~Sakmann,
\newblock ``{Regulation of synaptic efficacy by coincidence of postsynaptic APs
  and EPSPs},''
\newblock {\em Science}, vol. 275, no. 5297, pp. 213--215, 1997.

\bibitem{bi_poo}
G.-Q. Bi and M.-M. Poo,
\newblock ``Synaptic modifications in cultured hippocampal neurons: dependence
  on spike timing, synaptic strength, and postsynaptic cell type,''
\newblock {\em The Journal of Neuroscience}, vol. 18, no. 24, pp. 10464--10472,
  Dec. 1998.

\bibitem{zhang}
L.~Zhang, H.~Tao, C.~Holt, W.~Harris, and M.~M. Poo,
\newblock ``A critical window for cooperation and competition among developing
  retinotectal synapses,''
\newblock {\em Nature neuroscience}, vol. 395, pp. 37--44, Sep. 1998.

\bibitem{stdp_abbott}
L.~F. Abbott and S.~B. Nelson,
\newblock ``Synaptic plasticity: taming the beast,''
\newblock {\em Nature neuroscience}, vol. 3, pp. 1178--1183, Nov. 2000.

\bibitem{Bi_Wang_2002}
G.-Q. Bi and H.-X. Wang,
\newblock ``Temporal asymmetry in spike timing-dependent synaptic plasticity,''
\newblock {\em Physiology and Behavior}, vol. 77, pp. 551--555, 2002.

\bibitem{Froemke_Dan_2002}
R.~Froemke and Y.~Dan,
\newblock ``Spike-timing-dependent synaptic modification induced by natural
  spike trains,''
\newblock {\em Nature}, vol. 416, pp. 433--438, 2002.

\bibitem{wang}
H.~Wang, R.~Gerkin, D.~Nauen, and G.~Bi,
\newblock ``Coactivation and timing- dependent integration of synaptic
  potentiation and depression,''
\newblock {\em Nature neuroscience}, vol. 8, pp. 187--193, 2005.

\bibitem{Froemke2006}
R.~Froemke, I.~Tsay, M.~Raad, J.~Long, and Y.~Dan,
\newblock ``Contribution of individual spikes in burst-induced long-term
  synaptic modification,''
\newblock {\em Journal of Neurophysiology}, vol. 95, no. 3, pp. 1620--1629,
  2006.

\bibitem{SWAT}
J.J. Wade, L.J. McDaid, J.A. Santos, and H.M. Sayers,
\newblock ``{SWAT: A Spiking Neural Network Training Algorithm for
  Classification Problems},''
\newblock {\em IEEE Transactions on Neural Networks}, vol. 21, no. 11, pp.
  1817--1830, Nov. 2010.

\bibitem{Mitra2009}
S.~Mitra, S.~Fusi, and G.~Indiveri,
\newblock ``Real-time classification of complex patterns using spike-based
  learning in neuromorphic vlsi,''
\newblock {\em IEEE Transactions on Biomedical Circuits and Systems}, vol. 3,
  no. 1, pp. 32--42, Feb. 2009.

\bibitem{pfister_triplet}
J.-P. Pfister and W.~Gerstner,
\newblock ``Triplets of spikes in a model of spike timing-dependent
  plasticity,''
\newblock {\em The Journal of Neuroscience}, vol. 26, no. 38, pp. 9673--9682,
  Sep. 2006.

\bibitem{BCM}
E.~Bienenstock, L.~Cooper, and P.~Munro,
\newblock ``Theory for the development of neuron selectivity: orientation
  specificity and binocular interaction in visual cortex,''
\newblock {\em Journal of Neuroscience}, vol. 2, no. 1, pp. 32--48, Jan. 1982.

\bibitem{Izhikevich_lettercommunicated}
E.~Izhikevich and N.~Desai,
\newblock ``{Relating STDP to BCM},''
\newblock {\em Neural Computation}, vol. 15, no. 7, pp. 1511--1523, 2003.

\bibitem{Gjorgjieva}
J.~Gjorgjieva, C.~Clopath, J.~Audet, and J.-P. Pfister,
\newblock ``{A triplet spike-timing dependent plasticity model generalizes the
  Bienenstock-Cooper-Munro rule to higher-order spatiotemporal correlations},''
\newblock {\em Proceedings of the National Academy of Sciences}, vol. 108, no.
  48, pp. 19383--19388, 2011.

\bibitem{Mayr_algo}
C.~Mayr and J.~Partzsch,
\newblock ``Rate and pulse based plasticity governed by local synaptic state
  variables,''
\newblock {\em Frontiers in Computational Neuroscience}, vol. 2, no. 33, 2010.

\bibitem{Bo_Hasler}
J.~Hasler and B.~Marr,
\newblock ``Finding a roadmap to achieve large neuromorphic hardware systems,''
\newblock {\em Frontiers in neuroscience}, vol. 7, no. 00118, Sep. 2013.

\bibitem{arthur_gamma}
J.~Arthur and K.~Boahen,
\newblock ``Synchrony in silicon: The gamma rhythm,''
\newblock {\em IEEE Transactions on Neural Networks}, vol. 18, no. 6, pp.
  1815--1825, Nov. 2007.

\bibitem{giacomo_spikestdp}
G.~Indiveri, E.~Chicca, and R.~Douglas,
\newblock ``{A VLSI array of low-power spiking neurons and bistable synapses
  with spike-timing dependent plasticity},''
\newblock {\em IEEE Transactions on Neural Networks}, vol. 17, no. 1, pp.
  211--221, Jan. 2006.

\bibitem{dudek-2012}
J.~Wijekoon and P.~Dudek,
\newblock ``{VLSI circuits implementing computational models of neocortical
  circuits},''
\newblock {\em Journal of Neuroscience Methods}, vol. 210, no. 1, pp. 93--109,
  2012.

\bibitem{Koickal_Tara}
T.~J. Koickal, A.~Hamilton, T.~S. Lim, J.~A. Covington, J.~W. Gardner, and
  T.~C. Pearce,
\newblock ``{Analog VLSI circuit implementation of an adaptive neuromorphic
  olfaction chip},''
\newblock {\em IEEE Transactions on Circuits and Systems I: Regular Papers},
  vol. 54, no. 1, pp. 60--73, Jan. 2007.

\bibitem{STDP_Mems}
B.~Linares-Barranco, T.~Serrano-Gotarredona, A.~L. Camuñas-Mesa, A.~J.
  Perez-Carrasco, C.~Zamarreño-Ramos, and T.~Masquelier,
\newblock ``On spike-timing-dependent-plasticity, memristive devices, and
  building a self-learning visual cortex,''
\newblock {\em Frontiers in Neuroscience}, vol. 5, no. 26, 2011.

\bibitem{mostafa_triple_2013}
M.~R. Azghadi, S.~Al-Sarawi, D.~Abbott, and N.~Iannella,
\newblock ``A neuromorphic vlsi design for spike timing and rate based synaptic
  plasticity,''
\newblock {\em Neural Networks}, vol. 45, pp. 70--82, 2013.

\bibitem{brink_learningfg}
S.~Brink, S.~Nease, P.~Hasler, S.~Ramakrishnan, R.~Wunderlich, A.~Basu, and
  B.~Degnan,
\newblock ``A learning-enabled neuron array ic based upon transistor channel
  models of biological phenomenon,''
\newblock {\em IEEE Transactions on Biomedical Circuits and Systems}, vol. 7,
  no. 1, pp. 71--81, Feb. 2013.

\bibitem{FG_STDP_Pankaala}
M.~Pankaala, M.~Laiho, and P.~Hasler,
\newblock ``{Compact floating-gate learning array with STDP},''
\newblock in {\em Proceedings of the International Joint Conference on Neural
  Networks}, Atlanta, GA, Jun. 2009, pp. 2409--2415.

\bibitem{Roshan_IJCNN2014}
R.~Gopalakrishnan and A.~Basu,
\newblock ``{Robust doublet STDP in a floating-gate synapse},''
\newblock in {\em Proceedings of the International Joint Conference on Neural
  Networks}, Beijing, China, Jul. 2014, pp. 4296--4301.

\bibitem{Roshan_TNNLS2015}
R.~Gopalakrishnan and A.~Basu,
\newblock ``{On the Non-STDP Behavior and Its Remedy in a Floating-Gate
  Synapse},''
\newblock {\em IEEE Transactions on Neural Networks and Learning Systems}, vol.
  26, no. 99, pp. 2596--2601, Feb. 2015.

\bibitem{haas}
A.~M. Haas,
\newblock ``{Compact circuits and adaptation techniques for implementing
  adaptive neurons and synapses with spike timing dependent plasticity
  (STDP)},''
\newblock {\em Patent US20090292661}, Nov.~26, 2009.

\bibitem{liu_iscas_syn}
S.-C. Liu and R.~Mockel,
\newblock ``{Temporally learning floating-gate VLSI synapses},''
\newblock in {\em Proceedings of the International Symposium on Circuits and
  Systems}, Seattle, WA, May. 2008, pp. 2154--57.

\bibitem{Smith2014}
A.W. Smith, L.J. McDaid, and S.~Hall,
\newblock ``A compact spike-timing-dependent-plasticity circuit for floating
  gate weight implementation,''
\newblock {\em Neurocomputing}, vol. 124, pp. 210 -- 217, 2014.

\bibitem{fg_stdp}
S.~Ramakrishnan, P.~Hasler, and C.~Gordon.,
\newblock ``Floating gate synapses with spike-time-dependent plasticity,''
\newblock {\em IEEE Transactions on Biomedical Circuits and Systems}, vol. 5,
  no. 3, pp. 244--252, Jun. 2011.

\bibitem{Roshan_ISCAS2015}
R.~Gopalakrishnan and A.~Basu,
\newblock ``Triplet spike time dependent plasticity in a floating-gate
  synapse,''
\newblock in {\em Proceedings of the International Symposium on Circuits and
  Systems}, Lisbon, Portugal, May. 2015, pp. 710--713.

\bibitem{sjostrom}
P.~Sjostrom, G.~Turrigiano, and S.~Nelson,
\newblock ``Rate, timing, and cooperativity jointly determine cortical synaptic
  plasticity,''
\newblock {\em Neuron}, vol. 32, no. 6, pp. 1149--1164, Dec. 2001.

\bibitem{basu_iscas}
P.~Hasler, A.~Basu, and S.~Koziol,
\newblock ``Above threshold pfet injection modeling intended for programming
  floating-gate systems,''
\newblock in {\em Proceedings of the International Symposium on Circuits and
  Systems}, New Orleans, LA, USA, May. 2007, pp. 1557--1560.

\bibitem{Bamford2012}
S.A. Bamford, A.F. Murray, and D.J. Willshaw,
\newblock ``Spike-timing-dependent plasticity with weight dependence evoked
  from physical constraints,''
\newblock {\em IEEE Transactions on Biomedical Circuits and Systems}, vol. 6,
  no. 4, pp. 385--398, Aug. 2012.

\bibitem{Tanaka2009}
H.~Tanaka, T.~Morie, and K.~Aihara,
\newblock ``{A CMOS spiking neural network circuit with symmetric/asymmetric
  STDP function},''
\newblock {\em IEICE Transactions on Fundamentals of Electronics Communications
  and Computer Sciences}, vol. E92-A, no. 7, pp. 1690--1698, Jul. 2009.

\bibitem{Schemmel2006}
J.~Schemmel, A.~Grubl, K.~Meier, and E.~Mueller,
\newblock ``{Implementing synaptic plasticity in a VLSI spiking neural network
  model},''
\newblock in {\em Proceedings of the International Joint Conference on Neural
  Networks}, Vancouver, BC, Jul. 2006, pp. 1--6.

\bibitem{Mostafa2014}
M.R. Azghadi, N.~Iannella, S.F. Al-Sarawi, G.~Indiveri, and D.~Abbott,
\newblock ``Spike-based synaptic plasticity in silicon: Design, implementation,
  application, and challenges,''
\newblock {\em Proceedings of the IEEE}, vol. 102, no. 5, pp. 717--737, May.
  2014.

\bibitem{WangZQ2012}
Z.Q. Wang, H.Y. Xu, X.H. Li, H.~Yu, Y.C. Liu, and X.J. Zhu,
\newblock ``Synaptic learning and memory functions achieved using oxygen ion
  migration/diffusion in an amorphous ingazno memristor,''
\newblock {\em Advanced Functional Materials}, vol. 22, no. 13, pp. 2759--2765,
  2012.

\bibitem{Huajin_NECO_2013}
J.~Hu, H.~Tang, K.C. Tan, H.~Li, and L.~Shi,
\newblock ``A spike-timing-based integrated model for pattern recognition,''
\newblock {\em Neural Computation}, vol. 25, no. 2, pp. 450--472, Feb. 2013.

\bibitem{Huajin_TNNLS_2013}
Y.~Qiang, H.~Tang, K.C. Tan, and H.~Li,
\newblock ``Rapid feedforward computation by temporal encoding and learning
  with spiking neurons,''
\newblock {\em IEEE Transactions on Neural Networks and Learning Systems}, vol.
  24, no. 10, pp. 1539--1552, Oct. 2013.

\bibitem{giacomo_cognition}
E.~Chicca, F.~Stefanini, C.~Bartolozzi, and G.~Indiveri,
\newblock ``{Neuromorphic Electronic Circuits for Building Autonomous Cognitive
  Systems},''
\newblock {\em Proceedings of the IEEE}, vol. 102, no. 9, pp. 1367--1387, Sep.
  2014.

\bibitem{neurogrid}
B.~V. Benjamin, P.~Gao, E.~McQuinn, S.~Choudhary, A.~R. Chandrasekaran, J-M.
  Bussat, R.~Alvarez-Icaza, J.~V. Arthur, P.~A. Merolla, and K.~Boahen,
\newblock ``Neurogrid: A mixed-analog-digital multichip system for large-scale
  neural simulations,''
\newblock {\em Proceedings of the IEEE}, vol. 48, no. 8, pp. 1943--1953, Aug.
  2014.

\bibitem{spinnaker-jssc}
E.~Painkras, L.~A. Plana, J.~Garside, S.~Temple, F.~Galluppi, C.~Patterson,
  D.~R. Lester, A.~D. Brown, and S.~B. Furber,
\newblock ``Spinnaker: A 1-w 18-core system-on-chip for massively-parallel
  neural network simulation,''
\newblock {\em IEEE Journal of Solid-State Circuits}, vol. 102, no. 5, pp.
  699--716, Apr. 2013.

\bibitem{mead-neuromorphic}
C.~Mead,
\newblock ``Neuromorphic electronic systems,''
\newblock {\em Proceedings of the IEEE}, vol. 78, pp. 1629--1636, Oct. 1990.

\end{thebibliography}
\end{document}